\documentclass[10pt,twocolumn,letterpaper]{article}

\usepackage{iccv}
\usepackage{times}
\usepackage{epsfig}
\usepackage{graphicx}
\usepackage{amsmath}
\usepackage{amssymb}

\usepackage{caption}
\usepackage{subfigure}
\usepackage{color}
\usepackage{multicol}
\usepackage{multirow}
\usepackage{makecell}
\usepackage{pstricks} 
\usepackage{authblk}
\usepackage{booktabs}

\def\NumberDown#1#2{$#1^{\textcolor{green}{\downarrow}#2}$}
\def\NumberUp#1#2{$#1^{\textcolor{red}{\uparrow}#2}$}

\def\NumberUpBf#1#2{$\bf{#1^{\textcolor{red}{\uparrow}#2}}$}
\definecolor{darkgreen}{RGB}{153,196,131}
\definecolor{darkred}{RGB}{237,102,109}

\usepackage[breaklinks=true,bookmarks=false]{hyperref}

\iccvfinalcopy 

\def\iccvPaperID{5569} 

\ificcvfinal\fi

\begin{document}

\title{MT-ORL: Multi-Task Occlusion Relationship Learning}

\author{Panhe~Feng $^{1,2}$ \quad Qi~She $^{2}$ \quad Lei~Zhu $^{1}$ \quad Jiaxin~Li $^2$ \quad Lin~ZHANG $^2$ \quad Zijian~Feng $^2$ \quad Changhu~Wang $^{2}$ \quad Chunpeng~Li $^{1}$ \quad Xuejing~Kang \thanks{corresponding author.}$^{*,1}$ \quad Anlong~Ming $^{1}$
	\vspace{-1.5mm}\\
	$^1$Beijing University of Posts and Telecommunications. \quad $^2 $ByteDance Inc.
	\vspace{1.5mm}\\
	{\tt\small \{fengpanhe,sheqi1991\}@gmail.com,zhulei@stu.pku.edu.cn,\{lijx1992,1999forrestz,vincent.fung13\}}
	{\tt\small @gmail.com,wangchanghu@bytedance.com,\{chunpeng.li,kangxuejing,mal\}@bupt.edu.cn}	
}

\maketitle
\ificcvfinal\thispagestyle{empty}\fi

\begin{abstract}
	Retrieving occlusion relation among objects in a single image is challenging due to sparsity of boundaries in image. We observe two key issues in existing works: firstly, lack of an architecture which can exploit the limited amount of coupling in the decoder stage between the two subtasks, namely occlusion boundary extraction and occlusion orientation prediction, and secondly, improper representation of occlusion orientation. In this paper, we propose a novel architecture called \textbf{O}cclusion-shared and \textbf{P}ath-separated Network (\textbf{OPNet}), which solves the first issue by exploiting rich occlusion cues in shared high-level features and structured spatial information in task-specific low-level features. We then design a simple but effective orthogonal occlusion representation ({\textbf{OOR}}) to tackle the second issue. Our method surpasses the state-of-the-art methods by $6.1$\%/$8.3$\% Boundary-AP and $6.5$\%/$10$\% Orientation-AP on standard PIOD/BSDS ownership datasets. Code is available at \href{https://github.com/fengpanhe/MT-ORL}{https://github.com/fengpanhe/MT-ORL}.
\end{abstract}


\section{Introduction}
While the human visual system is capable of intuitively performing robust scene understanding and perception. The reasoning of occlusion relation is highly challenging for machines. Depending on the number, category, orientation, and position of objects, the boundaries of objects are elusive; no simple priors can be applied to recover the foreground and background in the scene. Occlusion relationship reasoning of objects from monocular images reveals relative depth differences among the objects in the scene, 
which is fundamental in computer vision applications, such as 
mobile robots \cite{marshall1996occlusion,sargin2009probabilistic,alatise2020review}, 
object detection \cite{gao2011segmentation, ayvaci2011detachable,tan2020efficientdet,pang2020multi,fan2020taking}, segmentation \cite{gao2011segmentation,zhang2015monocular, feldman2008motion,stringer2021cellpose,ke2021bcnet,zhu2017semantic,qi2019amodal},
monocular depth estimation/ordering \cite{qiu2020pixel,ramamonjisoa2019sharpnet,Ramamonjisoa_2020_CVPR,ren2006figure,maire2016affinity,zhan2020self}, 
and 3D reconstruction \cite{shan2014occluding,nikoohemat2020indoor,gonzalez2020nextmed}. 

Traditional methods \cite{hoiem2011recovering, teo2015fast, ren2006figure, saxena2006learning,jia2012learning} extract the occlusion boundary and infer occlusion relation by exploiting low-level visual cues with hand-crafted features, which is not effective due to the difficulty of defining boundaries and occlusion cues. Modern convolution neural networks (CNNs) \cite{wang2016doc,wang2018doobnet,lu2019occlusion} have significantly boosted the performance of occlusion relationship reasoning by commonly decomposing the task into two subtasks: occlusion boundary extraction and occlusion orientation prediction. The former aims to extract object boundaries from the image, while the latter targets are discovering the orientation relation. The occlusion relationship is then recovered by progressively accumulating the orientation information on the extracted boundaries. Despite the leap in occlusion relationship reasoning, we observe that two critical issues have rarely been discussed, which greatly impedes performances of previous work.

\begin{figure*}[ht]
	\begin{center}
		\subfigure[DOOBNet]{\includegraphics[scale=0.22]{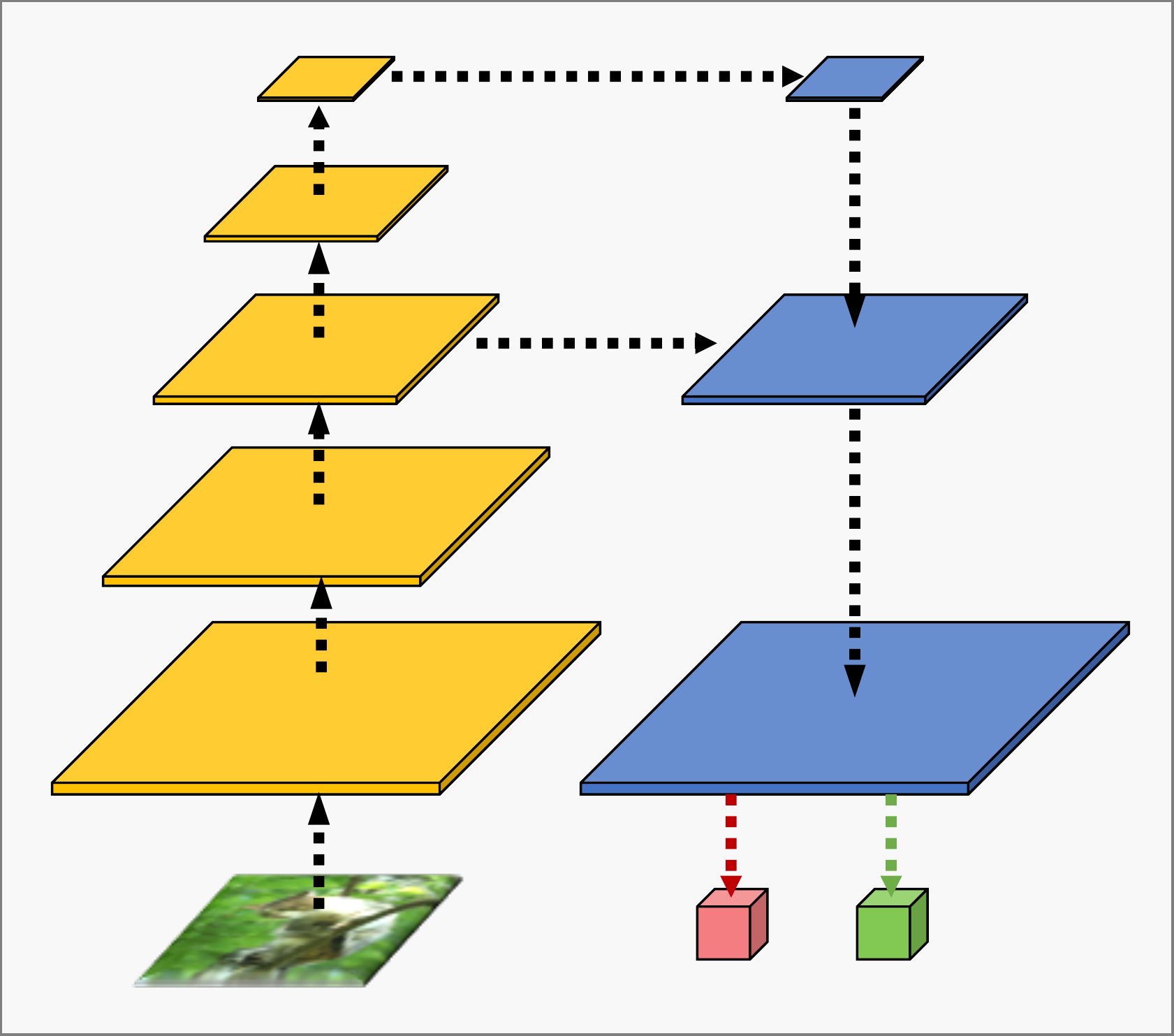}}
		\subfigure[OFNet]{\includegraphics[scale=0.22]{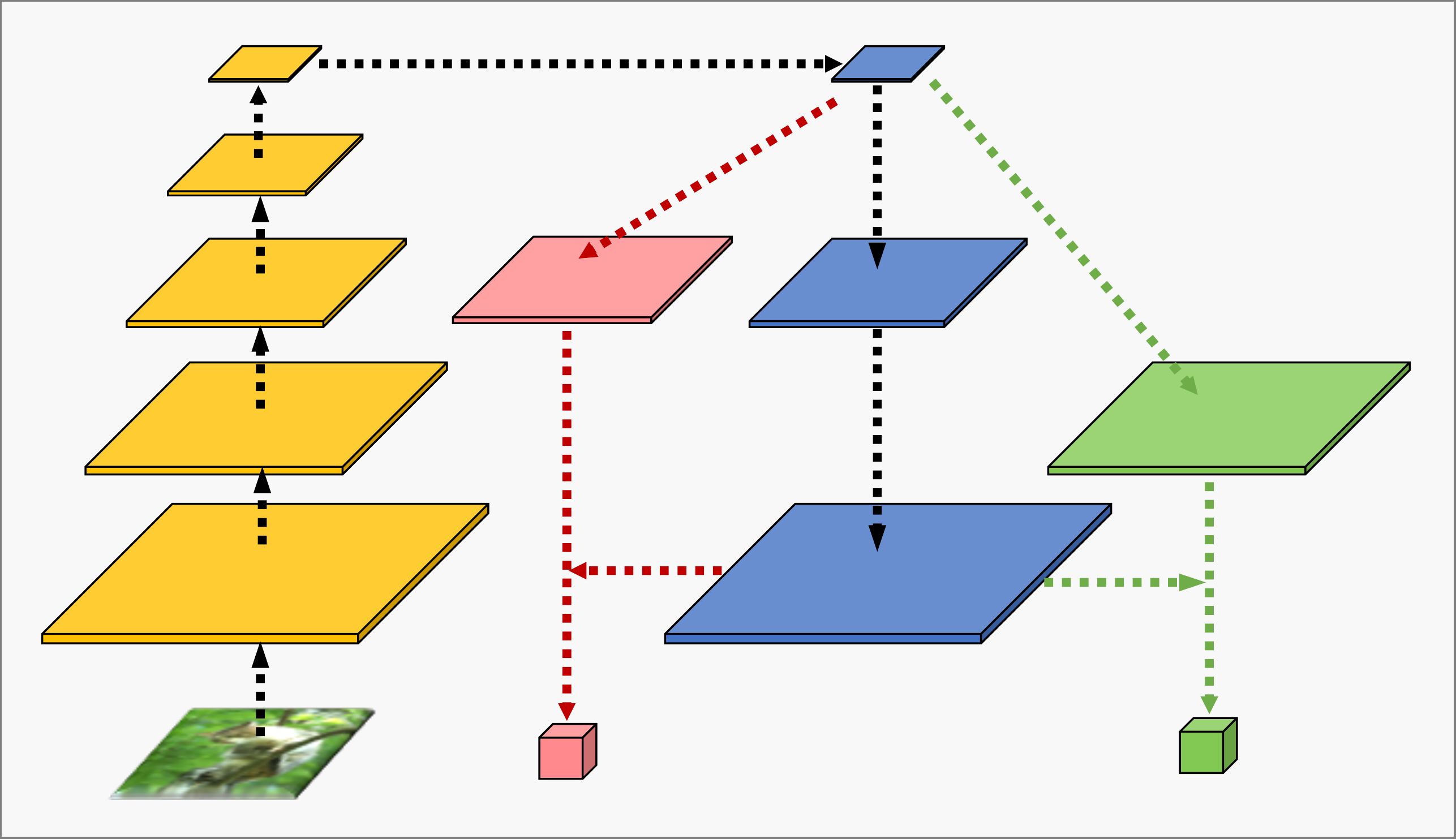}}
		\subfigure[Our OPNet]{\includegraphics[scale=0.22]{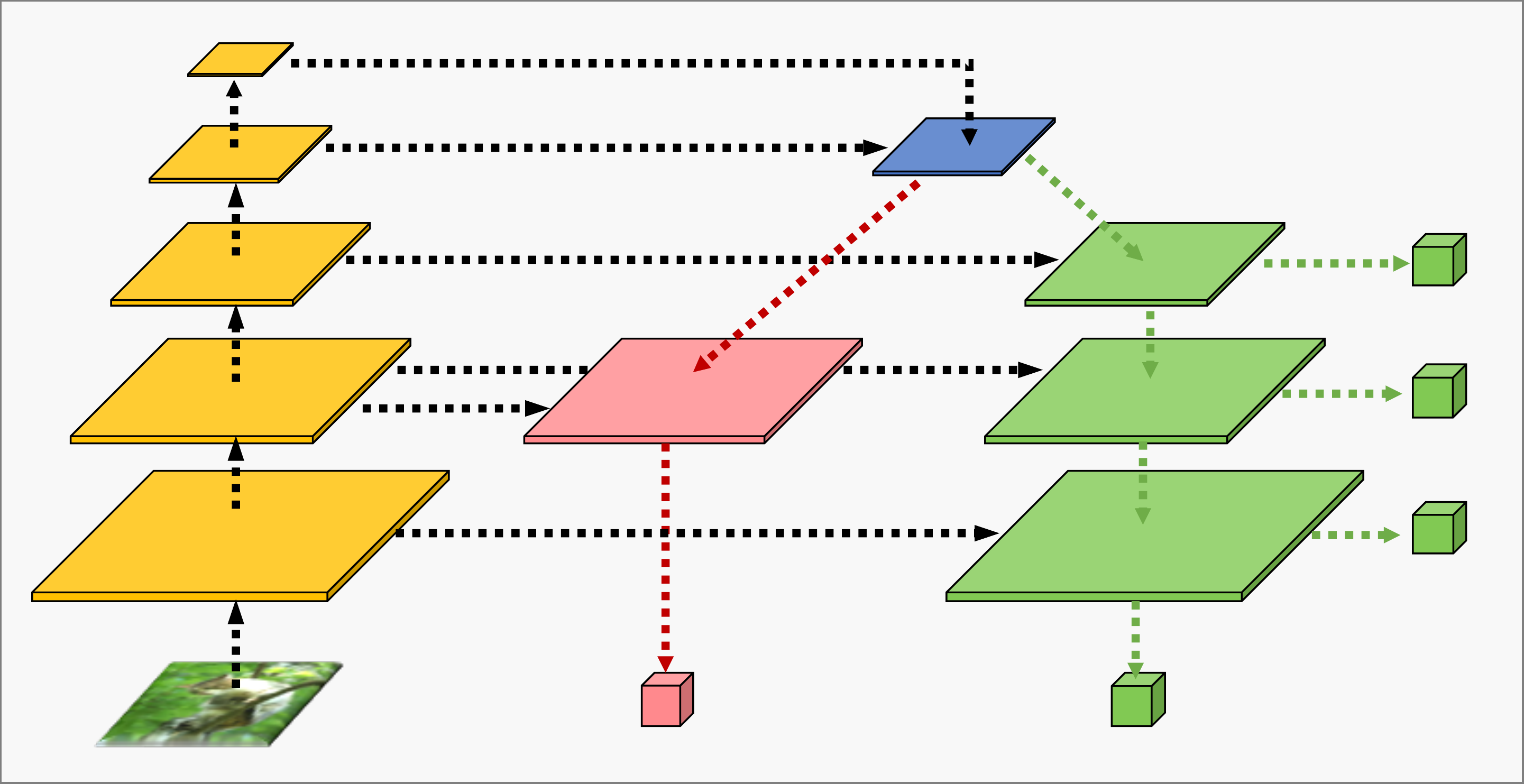}}
	\end{center}
	\vspace*{-8mm}
	\caption{Comparisons of different network architectures. \textcolor{yellow}{\bf{Yellow}} indicates encoder structure. \textcolor{green}{\bf{Green}} indicates boundary extraction path. \textcolor{pink}{\bf{Pink}} refers to occlusion orientation path. \textcolor{blue}{\bf{Blue}} indicates shared structure in decoder. Two color cubes indicate the boundary map and the orientation map of network output, respectively. Different from DOOBNet and OFNet, which over-shares local spatial information in larger feature maps, our OPNet only shares feature maps in deep stages while remaining path-separated in the shallow stage.
	}
	\vspace*{-4mm}
	\label{arch_compare}
\end{figure*}

The first issue is the lack of exploitation of the limited amount of coupling between the two subtasks. On one hand, shared global properties of the image and abstract high-level semantic features act as an effective initialization for the networks to learn low-level visual cues. Shared high-level features help different branches to generate consistent outputs under the same semantic guidance while different features may produce semantically misaligned results in the fusion stage. On the other hand, the two subtasks need to learn diverse and concrete properties in the lower stage with larger spatial size. Specifically, occlusion boundary extraction emphasizes more on locating boundaries, while occlusion orientation attaches more importance to exploiting relations between areas. In a word, the two subtasks are inherently coupled in abstract global features and decoupled in task-specific low-level features. However, previous architectures adopt a fully shared decoder~\cite{wang2018doobnet,lu2019occlusion}, resulting in an initialization feature map from the lower network stage.
																																
The second issue is the lack of a good representation of occlusion orientation. DOC~\cite{wang2016doc} first proposed to use pixel-wise continuous orientation variable as a representation, which however makes the design of loss function complex due to angle periodicity. DOOBNet~\cite{wang2018doobnet} then proposed to truncate the variable into range $(-\pi, \pi]$, which caused serious endpoint errors: angles close to $-\pi$ and $\pi$ produce large loss value while remaining relatively close orientation. This leaves the training process problematic. Therefore, designing a proper occlusion orientation representation remains challenging in the occlusion relationship reasoning.
																											
To tackle the first issue, we propose an Occlusion-shared and Path-separated Network (\textbf{OPNet}), which uses abstract feature maps with smaller spatial sizes for weight sharing and is then split into two separate task-specific paths, namely the occlusion boundary extraction path and occlusion orientation prediction path. Besides, multi-scale supervision \cite{hed2015} is adopted in the boundary extraction path to enhance the multi-scale features. To solve the second representation dilemma, we propose a simple and robust orthogonal occlusion representation ({\textbf{OOR}}), which represents orientation with two orthogonal vectors in horizontal and vertical directions, respectively.  Experiments show our method is effective and boosts the performances of both object boundary extraction and occlusion orientation prediction.

In a nutshell, our contributions are three-fold:
\begin{itemize} 
	\item We rethink the inherent properties of occlusion relationship reasoning, associated with two sub-tasks: occlusion boundary extraction and occlusion orientation regression. This motivates our advocate of a novel Occlusion-shared and Path-separated Network (OPNet) embodied with discrimination capability and expressiveness for visual occlusion reasoning as an alternative.
	\item We further propose the robust orthogonal occlusion representation(OOR) for predicting the occlusion orientation, which resolves the endpoint error and angle periodicity dilemma.
	\item The method works well on both PIOD \cite{wang2016doc} and BSDS ownership datasets \cite{Figure_Ground_Assignment_in_Natural_Images}, offering significantly better performances than the counterparts.
\end{itemize}

\begin{figure*}[ht]
	\begin{center}
		\subfigure[Overall architecture]{\includegraphics[width=0.85\linewidth]{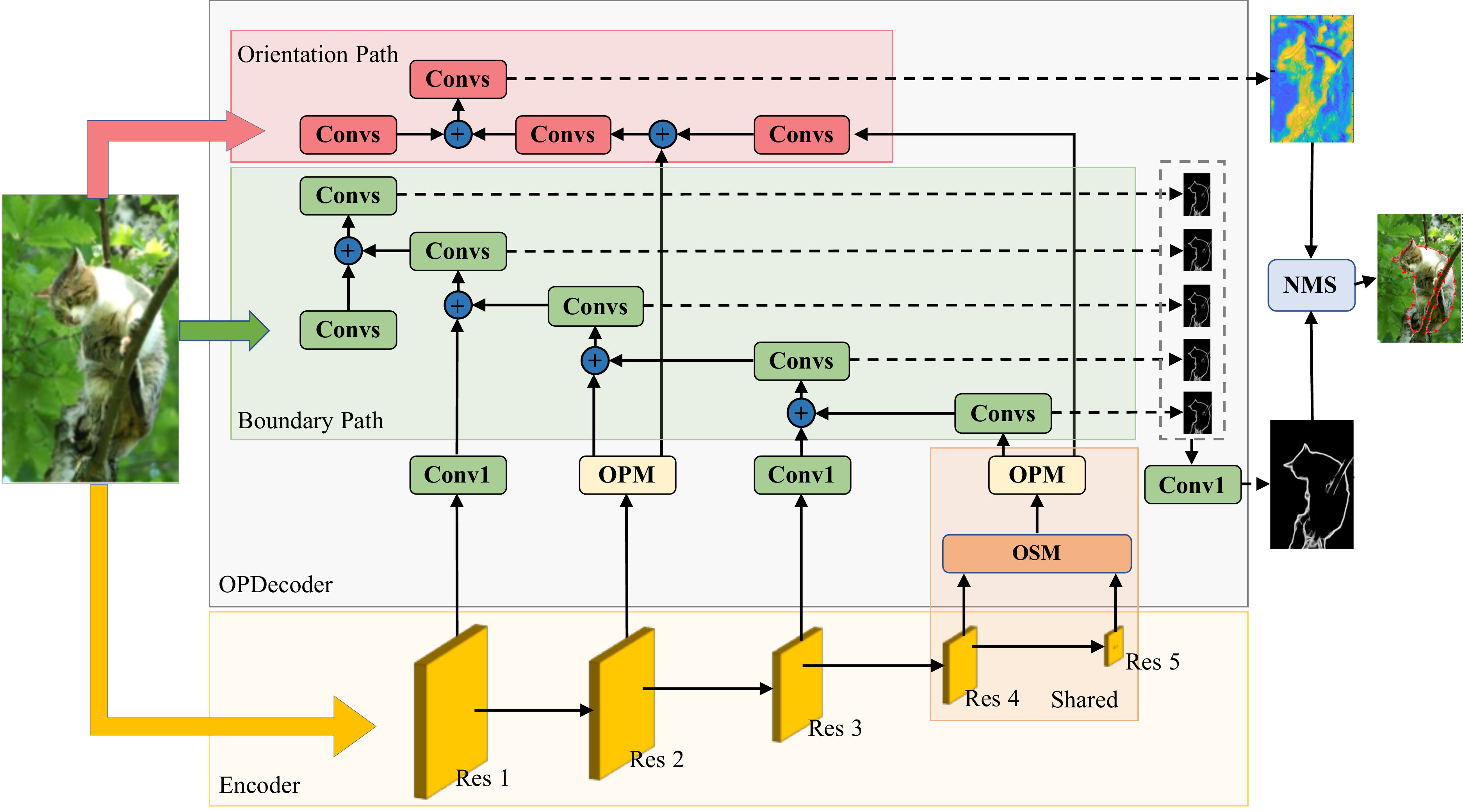}}
		\subfigure[ OPM: Orthogonal  Perception Module]{\includegraphics[width=0.4\linewidth]{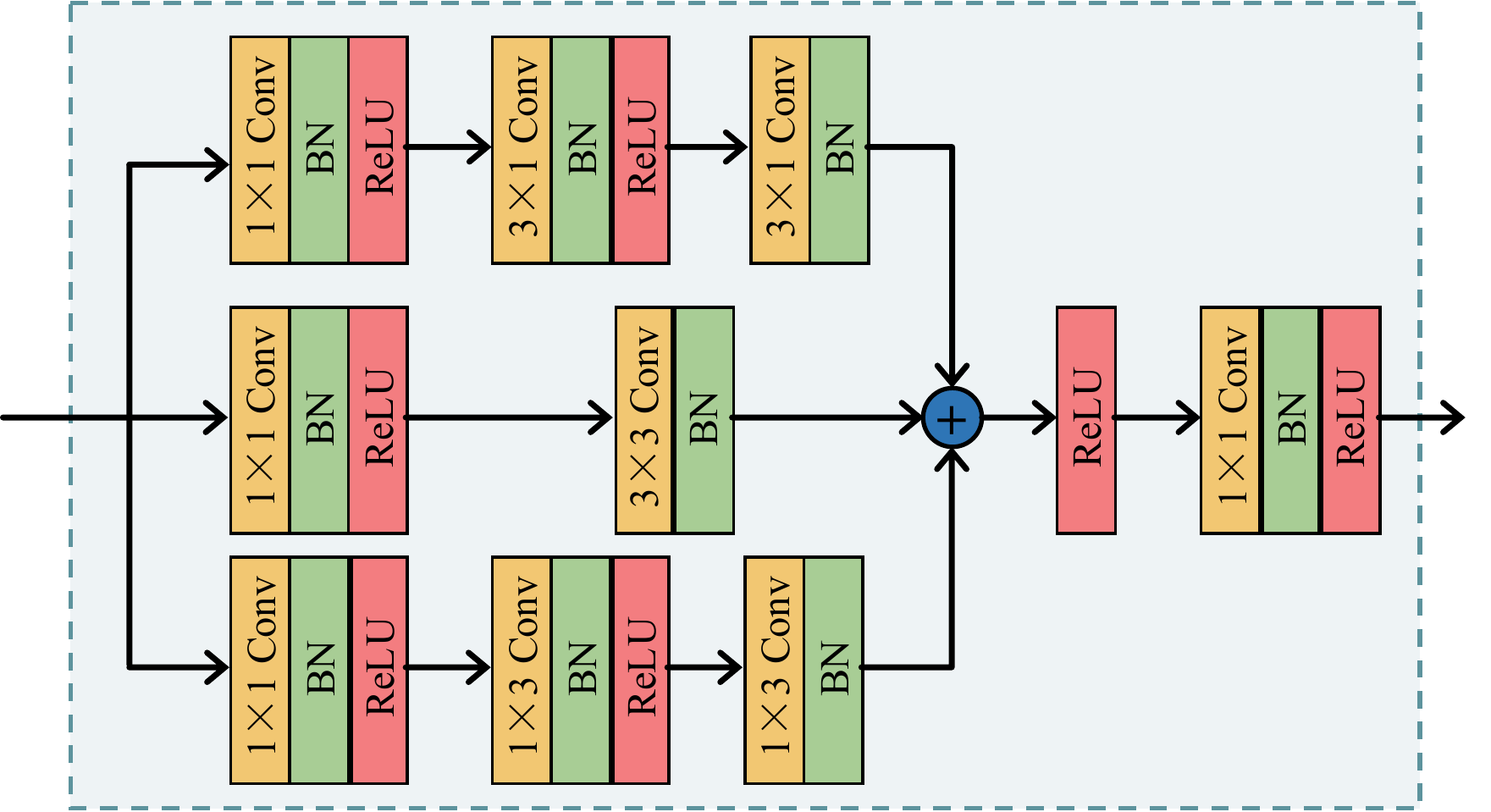}}
		\hspace{4em}
		\subfigure[ OSM: Occlusion-shared Module]{\includegraphics[width=0.25\linewidth]{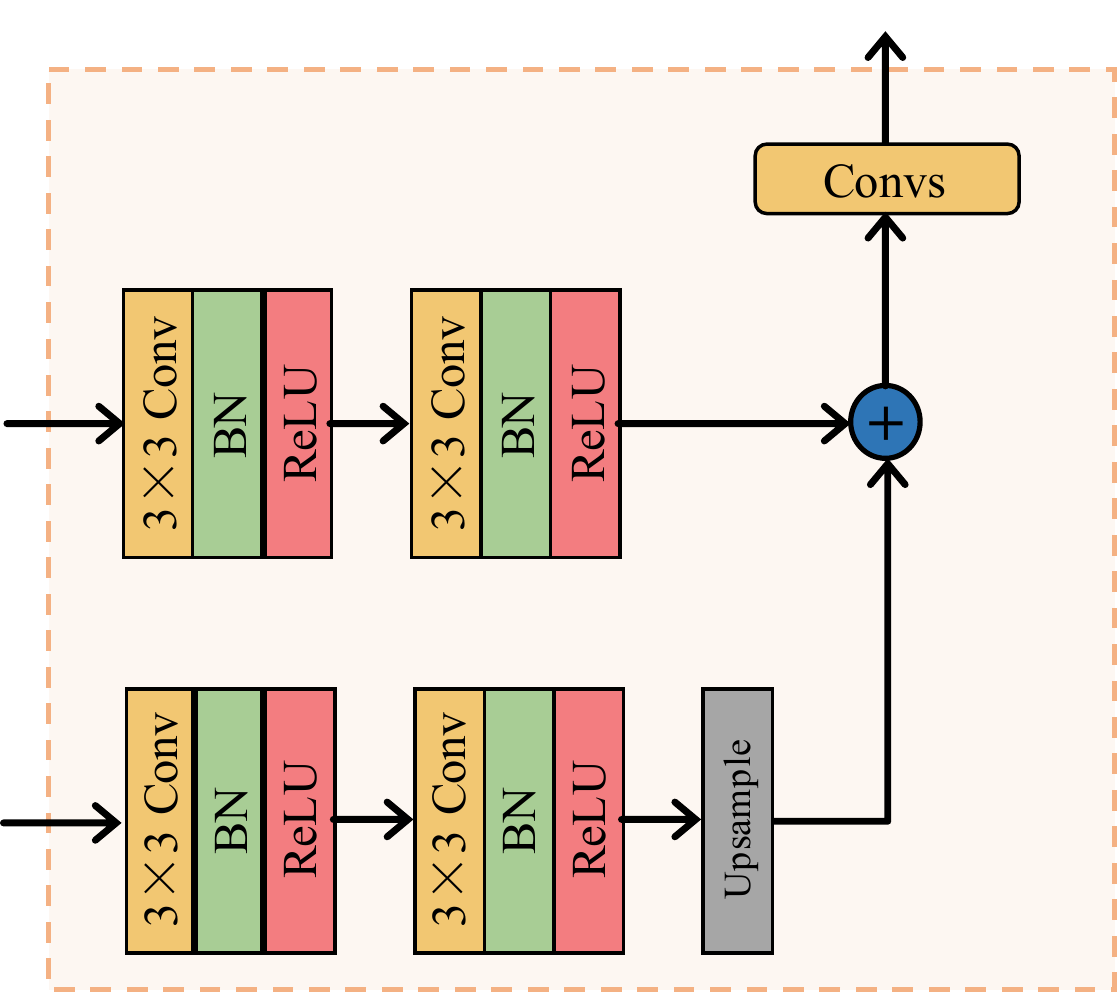}}
	\end{center}
	\vspace*{-6mm}
	\caption{The diagram of the proposed OPNet. 
		(a) Overall architecture. The input image is first encoded by an encoder and then input into the proposed decoder, which first aggregates shared features in the abstract high-level by OSM, and then generates boundary map and orientation map in two separated paths. NMS is used to fuse two outputs to produce the ultimate occlusion relationship map. (b) Orthogonal  Perception Module  (OPM). (c) Occlusion-shared Module (OSM).
		\textcircled{+} represents addition operation. 
		Details can be found in section \ref{sec_opnet}.
	}
	\vspace*{-2mm}
	\label{netstruce}
\end{figure*}
																																																																										
\section{Related Work}
Deep learning based occlusion relationship reasoning has achieved great success in recent years~\cite{wang2016doc,wang2018doobnet,lu2019occlusion}. DOC \cite{wang2016doc} represents occlusion relation by using a binary boundary indicator. It exploits local and non-local image cues to learn this representation and hence recovers the occlusion relation. DOOBNet \cite{wang2018doobnet} proposes to address extreme boundary/non-boundary class imbalance by up-weighting the loss contribution of false negative and false positive examples with a novel Attention Loss function. OFNet \cite{lu2019occlusion} builds on top of the encoder-decoder structure and side-output utilization. A fusion module is designed to precisely locate the object regions from the occlusion cues.

However, these methods either use two separate architectures for boundary extraction and orientation prediction or over-share local spatial information in shallow stages. Instead, our OPNet only shares features in deep stages of the decoder and adopts a multi-scale structure, as shown in Figure \ref{arch_compare}. We also propose an orthogonal orientation representation to tackle endpoint error and angle periodicity dilemma.

\section{The OPNet} \label{sec_opnet}

In this section, we detail our network architecture, Occlusion-shared and Path-separated Network (\textbf{OPNet}), which is composed of a shared encoder and our proposed novel Occlusion-shared and Path-separated decoder (\textbf{OPDecoder}). Following previous methods, we choose a ResNet50~\cite{resnet_2016,he2019bag} pretrained on ImageNet as our encoder. Our OPDecoder is equipped with Occlusion-shared Module (\textbf{OSM}),  Orthogonal Perception Module (\textbf{OPM}), and two separated paths, i.e, a boundary path and an orientation path. Extracted boundary map from boundary path and predicted orientation map from orientation path are then merged by Non-Maximum Suppression(NMS) to produce the final occlusion relationship map in the evaluation stage.

\subsection{The Occlusion-shared and Path-separated Decoder}

We define the deep stages in an encoder/decoder architecture as the ones where the feature map is deeper but spatially smaller, such as res4 or res5 layer in a ResNet architecture. Such deep stages contain less parameters and spatial details, thus are rich in abstract and global information that can be well adapted to downstream tasks. In contrast, shallow stages produce spatially-large feature maps with reduced channels. As a result, they are beneficial for learning task-specific low-level details. Our OPDecoder fully utilizes such distinct features at multiple stages.

\paragraph{Shared Deep Stage and Separated Shallow Stages}
Both boundary extraction and occlusion orientation prediction are dense prediction tasks aiming to recover pixel-wise spatial details with the understanding of high-level occlusion information. The former focuses on locating, while the latter expresses the relationship between regions where occlusion occurs. We thus propose our OPDecoder with a deep-to-shallow structure. Specifically, in order to better consider the connection and distinction between the two tasks, we divide the decoder into two parts, the shared deep stage with joint occlusion information and the separate shallow stages with concrete spatial details. The shared high-level semantic features in the deep stage learn more abstract occlusion cues by accepting supervision signals from both the boundary path and the orientation path. The boundary extraction path pays more attention to the spatial details, while the orientation regression path focuses on recovering regional relationships. Excessive sharing could result in mutual interference between the two tasks. Therefore, we also adopt separate structures in the shallow stages.

As shown in Figure~\ref{netstruce} (a), our decoder contains a deep occlusion-shared phase and a shallow spatial information branching phase. In the occlusion-shared phase, we use OSM to decode and aggregate two deepest stages' features and then pass them to the subsequent boundary path and orientation path.
In the shallow branching phase, considering the possible interference between the two tasks, we use OPM to enhance the features from shallow layers of the encoder, which will be fed to the following two paths. 

In the boundary extraction path, the shared feature map from OSM is progressively aggregated with transformed encoder features from deep to shallow stages. To better utilize spatial information at different scales, multi-scale boundary maps are generated, which are then fused together to extract a merged boundary map. Multi-scale supervision is applied to each generated boundary map.

Different from the boundary path, the orientation path is deliberately designed to only accept skip-connected features from the res2 layer to exploit less shared spatial details in shallow layers. It only keeps one supervision signal in the original scale due to the ambiguity of scaled orientation maps.

\begin{figure}[ht]
	\begin{center}
		\includegraphics[width=0.9\linewidth]{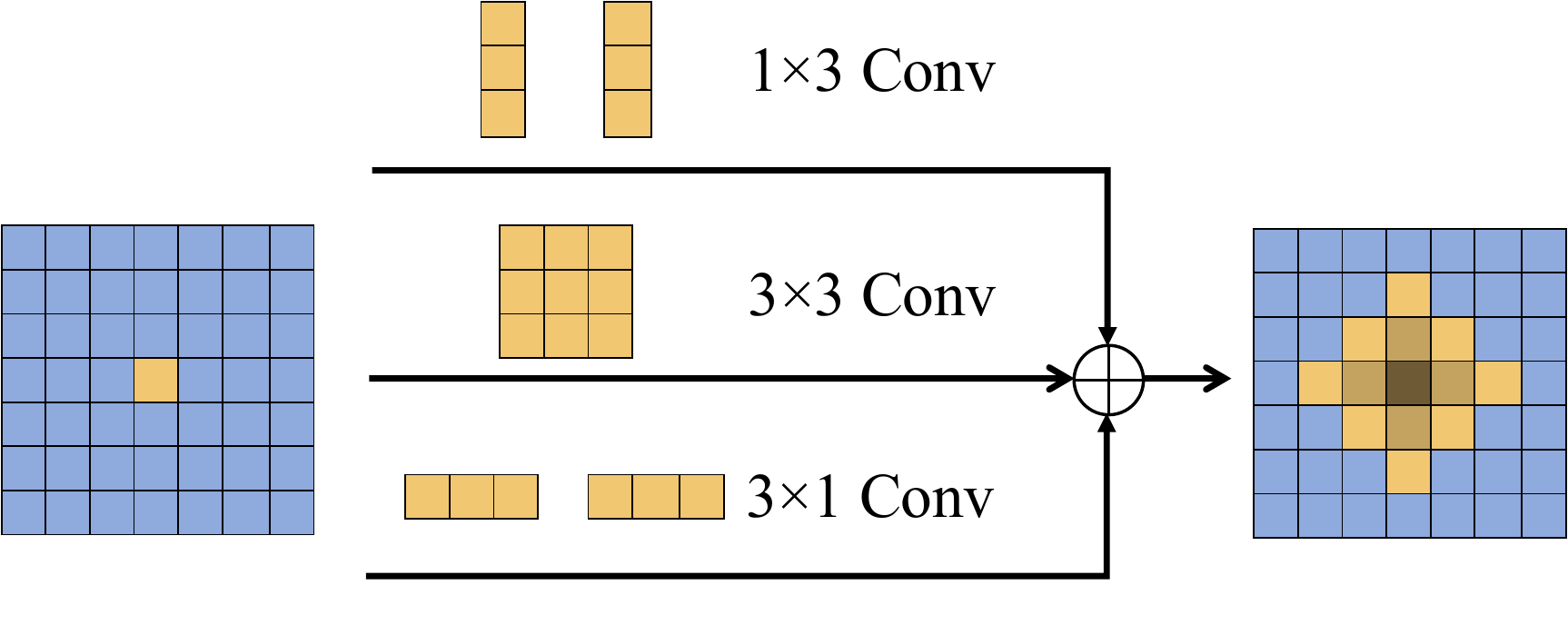}
	\end{center}
	\vspace*{-6mm}
	\caption{Receptive fields of the OPM. 
		Two consecutive convolutions of size $1 \times 3$ produce a receptive field of  $1 \times 5$ size in the vertical. And two convolutions of size $3 \times 1$ produce a receptive field of $5 \times 1$ size in the horizontal. Resulted convolution layers thus pay more attention along with the orthogonal directions, adhering to our orthogonal occlusion representation design.
	}
	\label{OPM1}
\end{figure}
																																																																						
\textbf{Orthogonal  Perception Module (OPM)}
As shown in Figure \ref{netstruce} (b), OPM contains three branches. The center branch consists of two $3 \times 3$ convolution blocks, while the other two are composed of two $1 \times 3$  and $3 \times 1$ strip convolution blocks, respectively.
Figure~\ref{OPM1} shows the receptive fields of module output features. 
The two stripe convolution paths simultaneously increase the receptive field and feature weight of the center pixel in orthogonal directions, and hence accumulate more informative orthogonal signals for further prediction of the OOR.
																																																																										
\textbf{Occlusion Shared Module (OSM)}
As shown in Figure \ref{netstruce} (c), OSM aggregates shared high-level features by fusing the upsampled res5 feature map and res4 feature map. It is designed to only merge features from deep stages, retaining abstract high-level shared occlusion features. OSM accepts supervision signals from both the boundary path and the orientation path to act on the high-level semantic features with larger receptive fields.

\subsection{Orthogonal Occlusion Representation} 
\begin{figure}[htbp!]
	\begin{center}
		\subfigure[]{\includegraphics[width=0.24\linewidth]{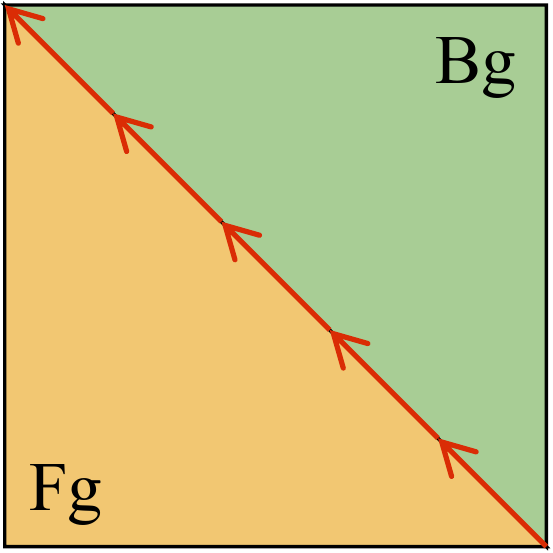}}
		\subfigure[]{\includegraphics[width=0.24\linewidth]{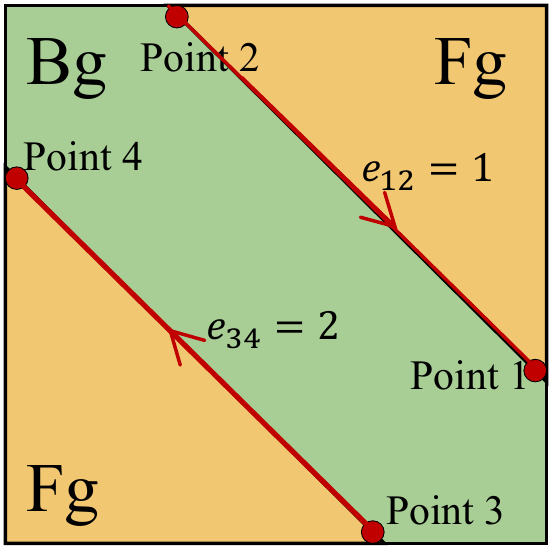}}
		\subfigure[]{\includegraphics[width=0.24\linewidth]{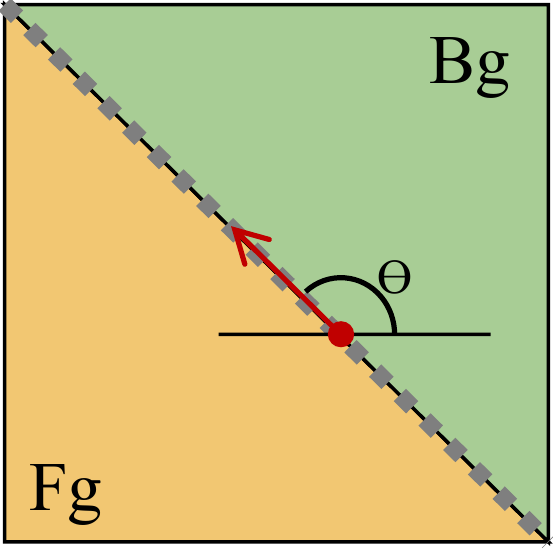}}
		\subfigure[]{\includegraphics[width=0.24\linewidth]{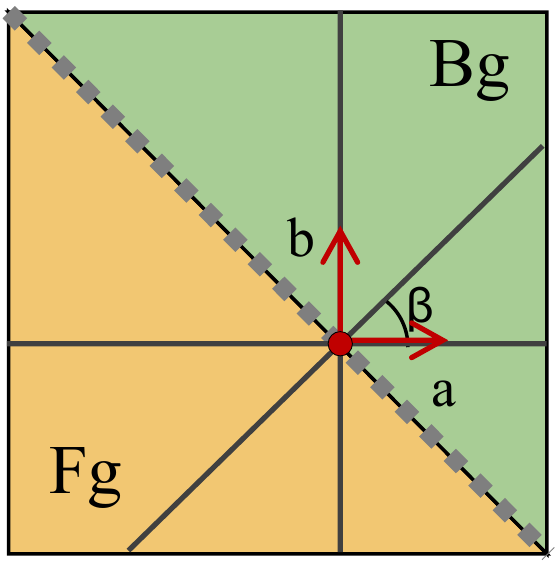}}
	\end{center}
	\vspace*{-6mm}
	\caption{The left rule and three ways to represent occlusion relations based on boundary. 
		(a) is the ``left" rule. (b) is the background label representation. (c) is the occlusion orientation representation. (d) is the orthogonal Occlusion representation.
	}
	\label{im_occ}
\end{figure}

In this section, we first point out the disadvantages of existing occlusion representations, i.e. angle periodicity and endpoint error. Different representation methods are then visualized for intuitive illustration. Finally, we present the novel orthogonal occlusion representation, which avoids the aforementioned two disadvantages by using two orthogonal vectors. Figure \ref{im_occ} shows a schematic diagram of four boundary-based representations of occlusion orientation, the lower left and upper right corners are the foreground and the background, respectively, while the diagonal line is the occlusion boundary between the two regions.

Figure \ref{im_occ} (a) shows the left rule of indicating foreground and background in an image. The occlusion boundary is represented with an arrow, the left-hand side of which is the foreground.
Figure \ref{im_occ} (b) is the label classification method based on complete boundaries commonly adopted by the traditional methods \cite{ming2015monocular,achanta2017superpixels,hoiem2011recovering,fowlkes2007local,arbelaez2010contour}, which classifies the obtained occlusion boundaries into two types. Label 1 indicates that the orientation of the boundary is from the starting point to the ending point, while label 2 means the opposite.  This method is considered ineffective due to the fact that detected boundaries are sparse and sometimes discontinuous.
Figure \ref{im_occ} (c) illustrates pixel-level orientation variable representation proposed in DOC \cite{wang2016doc}, which predicts boundary orientation by predicting a continuous orientation variable $\theta \in (-\pi, \pi]$ for each pixel in the image.
This method can well adapt to convolution blocks with dense prediction attributes and avoid the dependency on the quality of boundaries. It brings significant performance improvement on occlusion relationship reasoning and is widely adopted by CNN-based methods~\cite{wang2018doobnet,Lu2019ContextConstrainedAC,lu2019occlusion,wang2016doc}.

However, pixel-level orientation variable representation still has two critical issues. Firstly, it requires the network to regress an accurate angle for each boundary pixel, whereas predicting accurate angle variables brings an unnecessary burden. As long as the predicted angle has an error less than $ \pi / 2 $, the foreground-background relationship can be correctly recovered.
The second issue is the negative impact of angle periodicity on loss functions. The period of the angle variable is $ 2\pi $, meaning that two angles whose difference is a multiple of $2\pi$ are equivalent. However, such equivalence is hard to maintain in the loss function. In order to solve this problem, DOC~\cite{wang2016doc} defines a periodic segmentation function to represent the difference between two angles. But experiments show that this discontinuous loss function hinders the learning of the network.
DOOBNet adopts the strategy of truncating the predicted value to $ (-\pi, \pi] $, which however brings serious endpoint errors, i.e. angles near $-\pi $ and $ \pi $ could result in the higher loss while in fact staying close, introducing unnatural supervision signals near the endpoints.

We hereby propose our orthogonal occlusion representation (\textbf{OOR}): using a pair of orthogonal vectors, $ \vec{a} $ along the horizontal axis and $ \vec{b} $ along the vertical axis, to represent occlusion orientation. As shown in Figure \ref{im_occ}(d), both $ \vec{a} $ and $ \vec{b} $ point to the background near the occlusion boundary. OOR simplifies the prediction of occlusion orientation by instead pointing out the background both horizontally and vertically, greatly enhancing the robustness of prediction. Compared with previous the orientation variable representation, our OOR is simpler and works better, bypassing both angle periodicity and endpoint error dilemma.

\begin{figure}[ht]
	\begin{center}
		\includegraphics[width=0.9\linewidth]{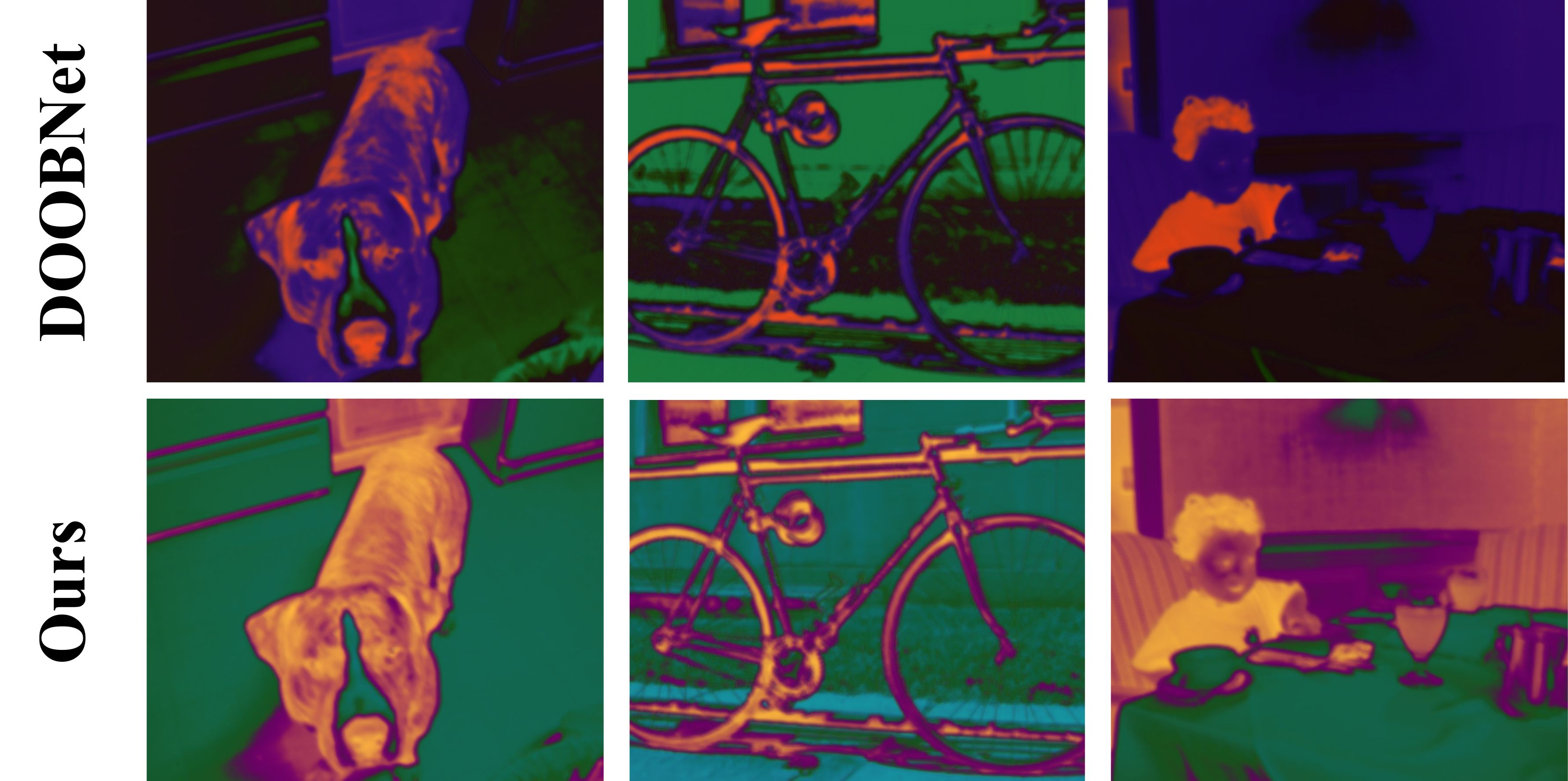}
	\end{center}
	\vspace*{-5mm}
	\caption{Visualization of the feature maps from networks trained with \textbf{different Orientation Representation}, which is produced by reducing feature dimension to $3$ via PCA.
		The first row is the result of using the DOOBNet\cite{wang2018doobnet} Orientation Representation, and the second row is Ours. Our result obviously shows clearer boundaries and occlusion relation, validating the effectiveness of the proposed OOR.}
	\label{occ_features}
\end{figure}

\begin{figure}[htbp]
	\centering
	\includegraphics[width=0.95\linewidth]{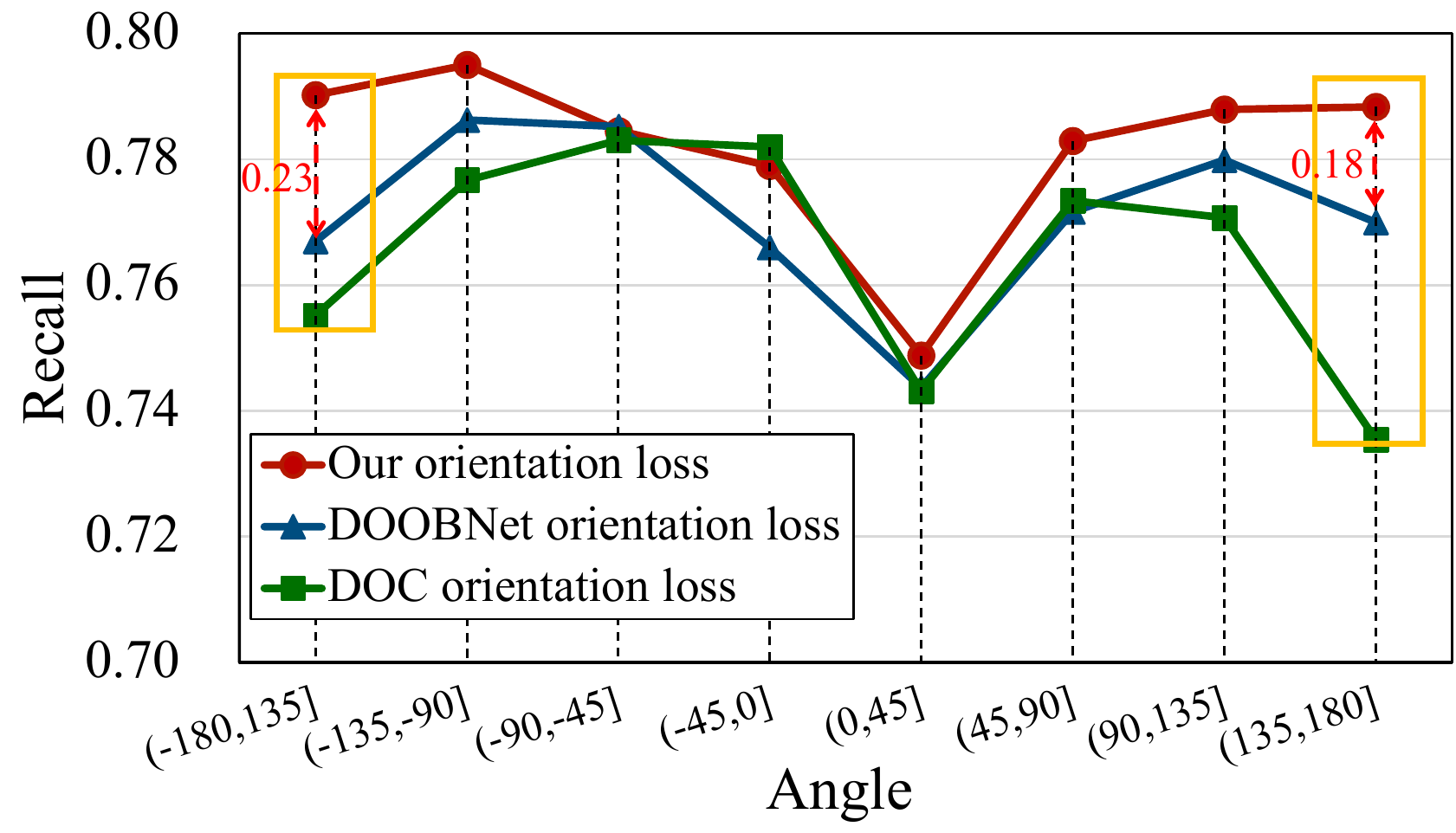}
	\caption{Recall vs. Angle curve on PIOD dataset. Angle is classified into eight bins and recall is calculated in each bin. Our proposed orientation representation shows obvious improvement near two endpoints $-\pi$ and $\pi$ (the yellow box).
	}
	\label{occ_bins}
	\vspace*{-4mm}
\end{figure}
																																																																										
Figure \ref{occ_features} visualizes the feature maps extracted from the Res2 layer trained with different orientation representations. Compared with the feature maps obtained by using DOOBNet's orientation representation and loss, the contrast between foreground and background of our result is more obvious. Experiments in Table.~\ref{res_table} show that our method improves all three occlusion-related metrics (O-ODS, O-ODS, O-AP) by at least $2\%$ upon orientation representation of DOOBNet, which shows stronger occlusion perception potential of our OOR.
Figure \ref{occ_bins} shows the comparisons of different methods on occlusion orientation recall towards different boundary directions. We divide $(-\pi, \pi]$ into eight bins and calculate the recall of orientation prediction in each bin. Results show that DOC's and DOOBNet's orientation representations and loss pay too much attention to the vertical direction and produce large errors near two endpoints due to the aforementioned endpoint error dilemma.

\subsection{Network Training}\label{net_train}
For an input image $I$, the ground truth label is represented as a pair of object boundary map and occlusion orientation map $\{Y, O\}$, where $Y = (y_j, j = 1, \dots ,|I|), y_j \in \{0, 1\}$. $O = (\theta_j, j = 1, \dots ,|I|), \theta_j \in (-\pi, \pi]$ and $\theta_j$ is invalid when $y_j=0$.  
In CNN-based methods of occlusion relationship reasoning,
occlusion boundary extraction and occlusion orientation prediction are trained simultaneously.
Thus the overall loss $\mathcal{L}$  is formulated as:
\begin{equation}
	\begin{aligned} \mathcal{L} = \mathcal{L}_{\text {B}}\end{aligned} + w_{\text{O}} \cdot \mathcal{L}_{\text{O}}, 
\end{equation}
where $w_{\text {O}}$ is the weight for occlusion orientation loss. $\mathcal{L}_{\text{B}}$ and  $\mathcal{L}_{\text {O}}$ represent the loss of boundary path and orientation path, respectively. In boundary path, we can get  a collection of five side boundary maps and a fused boundary map $\{\hat{Y}_{\text{s-1}}, \hat{Y}_{\text{s-2}}, \cdots, \hat{Y}_{\text{s-K}}, \hat{Y}_{\text{f}}\}$.
We formulate the loss of boundary extraction as, where in this work $K = 5$ : 
\begin{equation}
	\mathcal{L}_{\text {B}} = w_{\text{f}} \cdot \Gamma(\hat{Y}_{\text{f}}, Y) + 
	\sum_{i=1}^{K} w_{\text{s-i}} \cdot \Gamma(\hat{Y}_{\text{s-i}}, Y),
\end{equation}
where $w_{\text{s-i}}$ and $w_{\text{f}}$ are weights for the side boundary map loss and fused boundary map loss, respectively. $\hat{Y} = (y_j, j = 1, \dots ,|I|), y_j \in (0, 1)$   represents the predicted boundary map .
The function $\Gamma (\hat{Y}, Y)$ is computed with all pixels. Since the distribution of boundary/non-boundary pixels is heavily biased, we employ a class-balanced cross-entropy loss.
we define $Y_{-}$ and $Y_{+}$ as the non-boundary and boundary ground truth label sets in a mini-batch of images. We then define $\Gamma (\hat{Y}, Y)$ as:
\begin{align}
	\Gamma(\hat{Y}, Y) & = - \lambda \alpha \sum_{j \in Y_{-}} \log \left(1-\hat{y}_{j}\right)- (1-\alpha) \sum_{j \in Y_{+}} \log \left(\hat{y}_{j}\right). 
\end{align}
Here, $\alpha=\left|Y_{+}\right| /\left(\left|Y_{+}\right|+\left|Y_{-}\right|\right)$ can balance the boundary/non-boundary pixels, and $\lambda$ controls the weight of positive over negative samples.

The function $\mathcal{L}_{\text {O}}$ calculates the loss of the orientation only at the groundtruth boundary pixels.
Based on our representation, we proposed a novel loss called \textbf{Orthogonal Orientation Regression loss (OOR)}, which is defined as:
\begin{equation}
	\mathcal{L}_{\text{O}} = \text{OOR}(\hat{a}, \hat{b};\theta) = \operatorname{smooth}_{L_{1}}\Big(\Phi(\hat{a}, \hat{b};\theta)\Big), 
\end{equation}
in which we design the novel formulation as:
\begin{equation}
	\begin{split}
		\Phi(\hat{a}, \hat{b};\theta)=(\frac{\hat{a}}{\sqrt{\hat{a}^{2}+\hat{b}^2}}-\cos \theta_{j})^{2}
		+(\frac{\hat{b}}{\sqrt{\hat{a}^{2}+\hat{b}^{2}}}-\sin \theta_{j})^{2}.
	\end{split}
\end{equation}
${\hat{a}}$ and $\hat{b}$ represent prediction value of the network.
We define the $\vec{a} = (\hat{a}, 0)$ and $\vec{b} = (0, \hat{b})$. 
The $\theta$ is the ground truth of occlusion orientation.

\section{Experiment}
\begin{table*}[htbp]
	\renewcommand\arraystretch{1.1}
	\begin{center}
		\caption{Comparison between our method with others on PIOD dataset and BSDS ownership dataset. 
			$\dag$ indicates training boundary extraction branch only.
			$\ddag$ indicates training orientation prediction branch only.
			dor indicates using the orientation loss function proposed by DOC~\cite{wang2016doc}.
			dbr indicates the orientation loss function proposed by DOOBNet~\cite{wang2018doobnet}. 
			The red/green superscripts (${\textcolor{red}{^\uparrow}}$ / ${\textcolor{green}{^\downarrow}}$) indicate increasing/decreasing upon the state-of-the-art method (OFNet\cite{lu2019occlusion}). 
			All numbers are represented in percentage. Details of different evaluation metrics can be found in Section 4.1.
		}
		\label{res_table}
		\setlength{\tabcolsep}{0.2mm}{
			\scalebox{0.95}{
				\begin{tabular}
					{c c c c  c c c  | c c c c c c}
					\toprule[1.2pt]
					\multirow{3}{*}{\textbf{Method}}
					&\multicolumn{6}{c|}{\textbf{PIOD Dataset}}&\multicolumn{6}{c}{\textbf{BSDS ownership Dataset}}\cr
					\cmidrule(r){2-13} 
					               & \textbf{\makecell[c]{B-ODS}} & \textbf{B-OIS}         & \textbf{B-AP}          & \textbf{O-ODS}         & \textbf{O-OIS}         & \textbf{O-AP}          & \textbf{B-ODS}         & \textbf{B-OIS}         & \textbf{B-AP}          & \textbf{O-ODS}         & \textbf{O-OIS}         & \textbf{O-AP}\cr         
					\midrule[1pt]
					SRF-OCC        & $34.5$                       & $36.9$                 & $20.7$                 & $26.8$                 & $28.6$                 & $15.2$                 & $51.1$                 & $54.4$                 & $44.2$                 & $41.9$                 & $44.8$                 & $33.7$\cr                
					DOC-HED        & $50.9$                       & $53.2$                 & $46.8$                 & $46.0$                 & $47.9$                 & $40.5$                 & $65.8$                 & $68.5$                 & \underline{$60.2$}     & $52.2$                 & $54.5$                 & $42.8$\cr                
					DOC-DMLFOV     & $66.9$                       & $68.4$                 & $67.7$                 & $60.1$                 & $61.1$                 & $58.5$                 & $57.9$                 & $60.9$                 & $51.9$                 & $46.3$                 & $49.1$                 & $36.9$\cr                
					DOOBNet        & $73.6$                       & $74.6$                 & $72.3$                 & $70.2$                 & $71.2$                 & $68.3$                 & $64.7$                 & $66.8$                 & $53.9$                 & $55.5$                 & $57.0$                 & $44.0$\cr                
					\hline
					OFNet  $\dag$  & $73.9$                       & $75.0$                 & $68.5$                 & -                      & -                      & -                      & -                      & -                      & -                      & -                      & -                      & -\cr                     
					OFNet  $\ddag$ & -                            & -                      & -                      & $70.5$                 & $71.6$                 & $67.4$                 & -                      & -                      & -                      & -                      & -                      & -\cr                     
					OFNet &\underline{$75.1$} &\underline{$76.2$} & \underline{$77.0$} & \underline{$71.8$}& \underline{$72.8$} & \underline{$72.9$} & 
					\underline{$66.2$} & \underline{$68.9$} & $58.5$ & \underline{$58.3$} & \underline{$60.7$} & \underline{$50.1$}\cr
					\hline
					OPNet + dor &\NumberUp{76.6}{1.5} &\NumberUp{77.8}{1.6} &\NumberDown{76.1}{0.9} &\NumberUp{73.7}{1.9} &\NumberUp{74.6}{1.8} &\NumberDown{72.2}{0.7}
					&\NumberUp{66.2}{0} & \NumberDown{68.7}{0.2} & \NumberUp{63.2}{4.7} & \NumberDown{55.5}{2.8} & \NumberDown{57.5}{3.2} & \NumberDown{47.0}{3.1} \cr
					OPNet + dbr &\NumberUp{77.8}{2.7} & \NumberUp{79.0}{2.8} & \NumberUp{80.6}{3.6} & \NumberUp{74.7}{1.9} & \NumberUp{75.8}{3.0} & \NumberUp{75.9}{3.0}&
					\NumberUp{66.8}{0.6} & \NumberUp{68.9}{0} & \NumberUp{62.4}{3.9} &  \NumberDown{58.1}{0.2} & \NumberDown{60.0}{0.7} & \NumberUp{50.5}{0.4}\cr
					\hline
					Ours $\dag$    & \NumberUp{78.6}{3.5}         & \NumberUp{79.6}{3.4}   & \NumberUp{79.5}{2.5}   & -                      & -                      & -                      & -                      & -                      & -                      & -                      & -                      & -\cr                     
					Ours $\ddag$   & -                            & -                      & -                      & \NumberUp{76.1}{4.3}   & \NumberUp{77.0}{4.2}   & \NumberUp{76.1}{3.2}   & -                      & -                      & -                      & -                      & -                      & -\cr                     
					\bf{Ours}      & \NumberUpBf{79.5}{4.4}       & \NumberUpBf{80.4}{4.2} & \NumberUpBf{83.1}{6.1} & \NumberUpBf{77.1}{5.3} & \NumberUpBf{77.8}{5.0} & \NumberUpBf{79.4}{6.5} & \NumberUpBf{67.2}{1.0} & \NumberUpBf{70.3}{1.4} & \NumberUpBf{66.8}{8.3} & \NumberUpBf{61.9}{3.6} & \NumberUpBf{64.8}{3.4} & \NumberUpBf{60.1}{10}\cr 
					\bottomrule[0.8pt]
				\end{tabular}}
		}
		\label{result_table}
	\end{center}
	\vspace*{-6mm}
\end{table*}

\begin{figure*}[htbp]
	\begin{center}
		\subfigure[PRC of boundary  on PIOD.] {\includegraphics[width=0.23\linewidth]{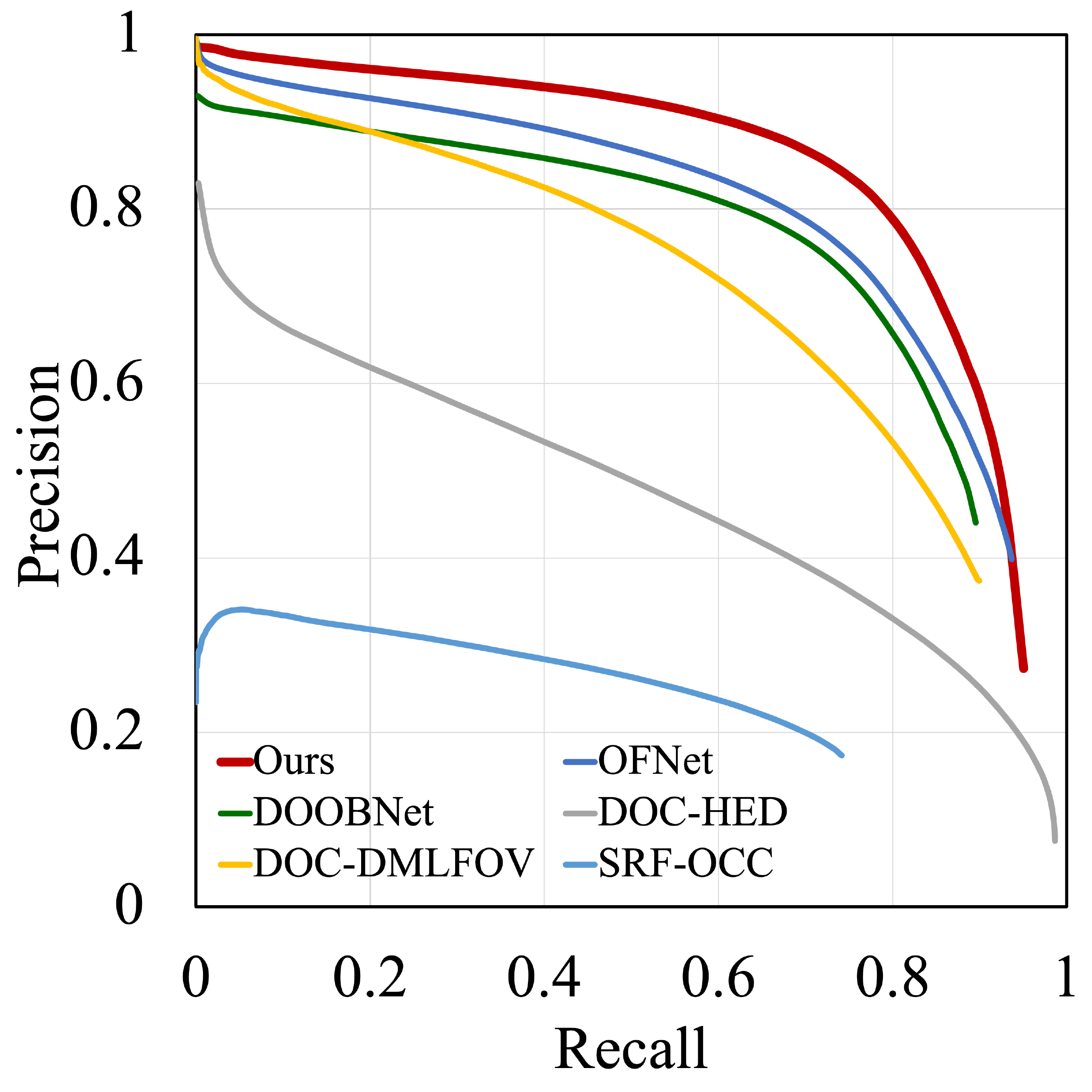} }
		\subfigure[PRC of occlusion  on PIOD.] {\includegraphics[width=0.23\linewidth]{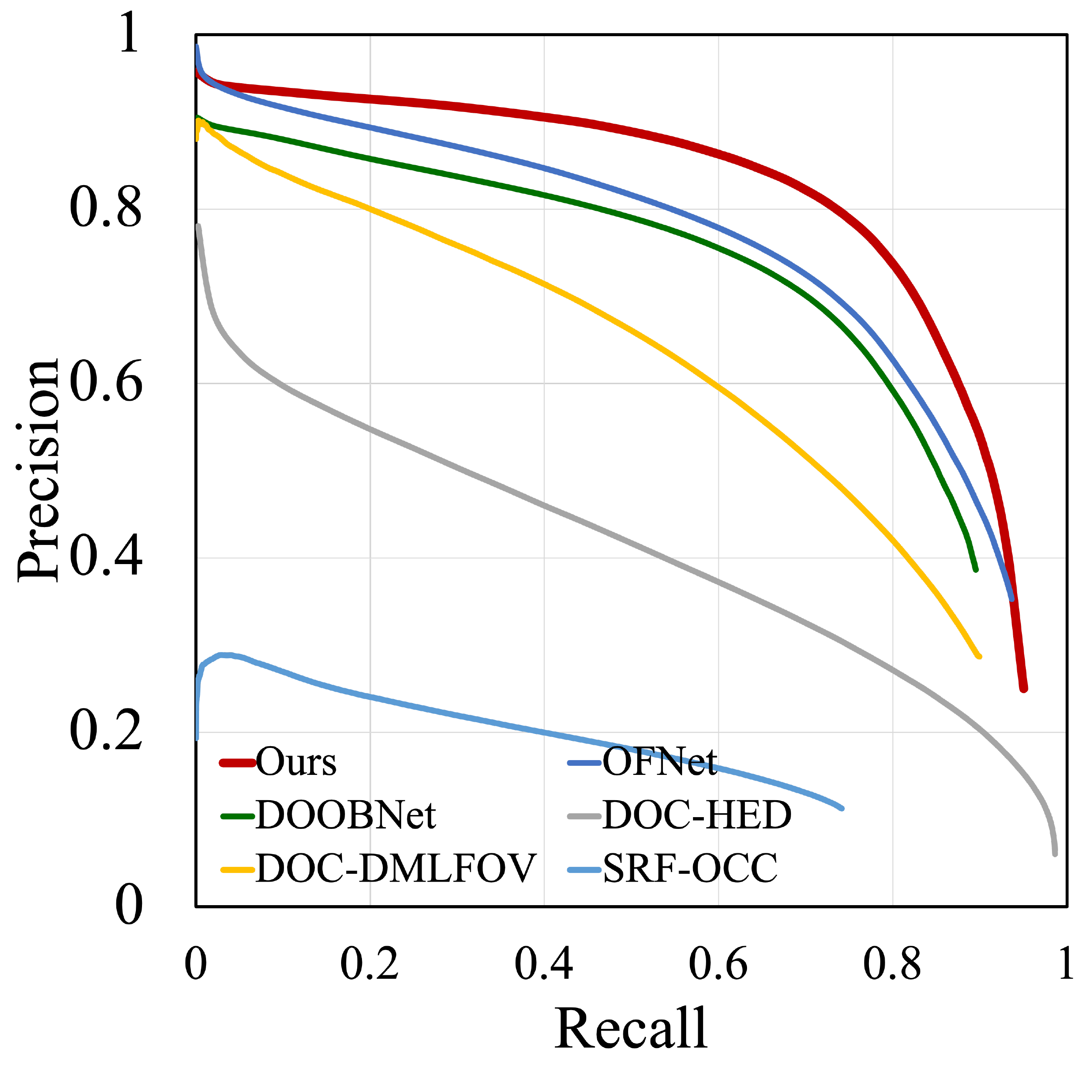} }
		\subfigure[PRC of boundary  on BSDS ownership.] {\includegraphics[width=0.23\linewidth]{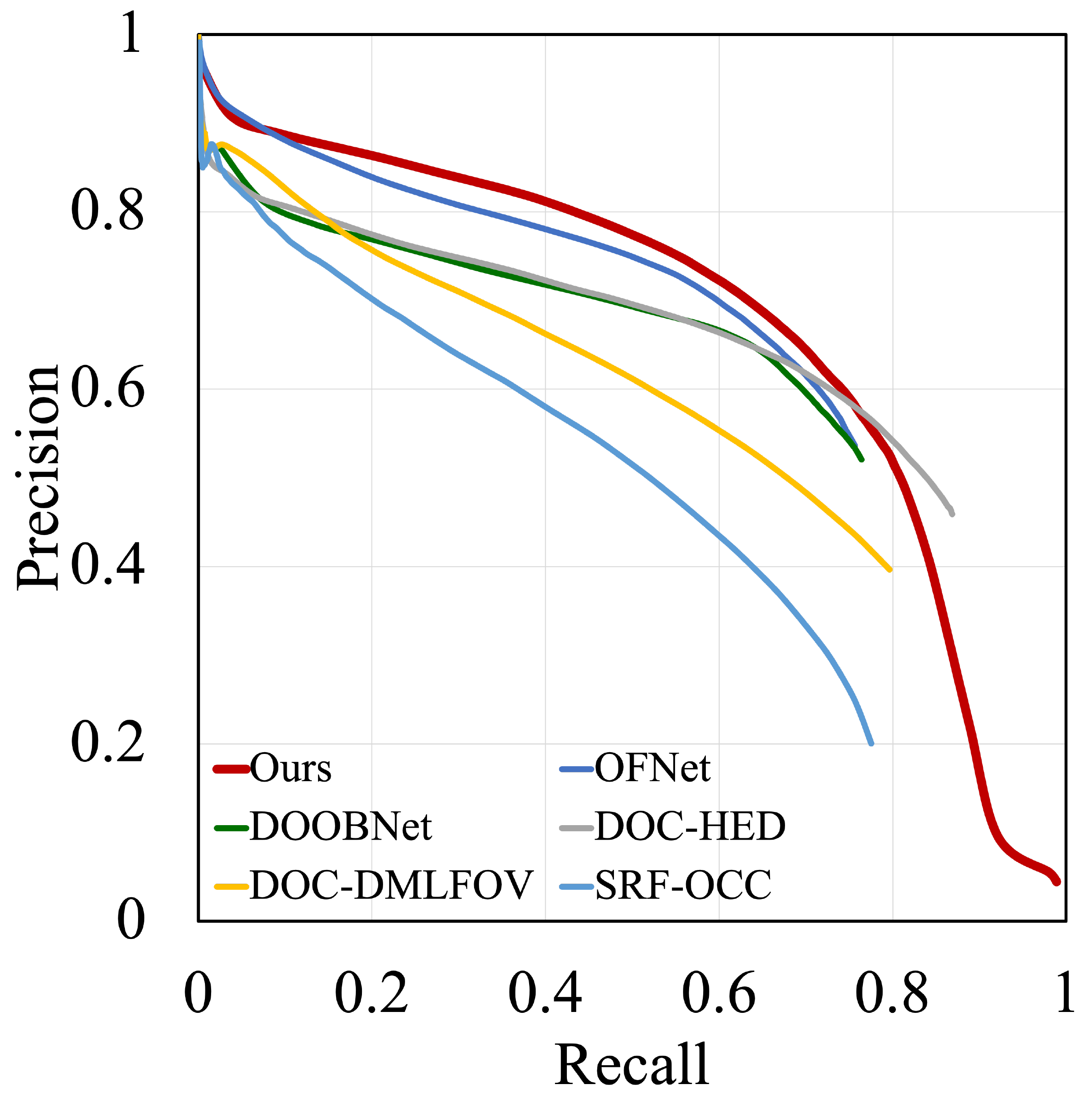} }
		\subfigure[PRC of occlusion  on BSDS ownership.] {\includegraphics[width=0.23\linewidth]{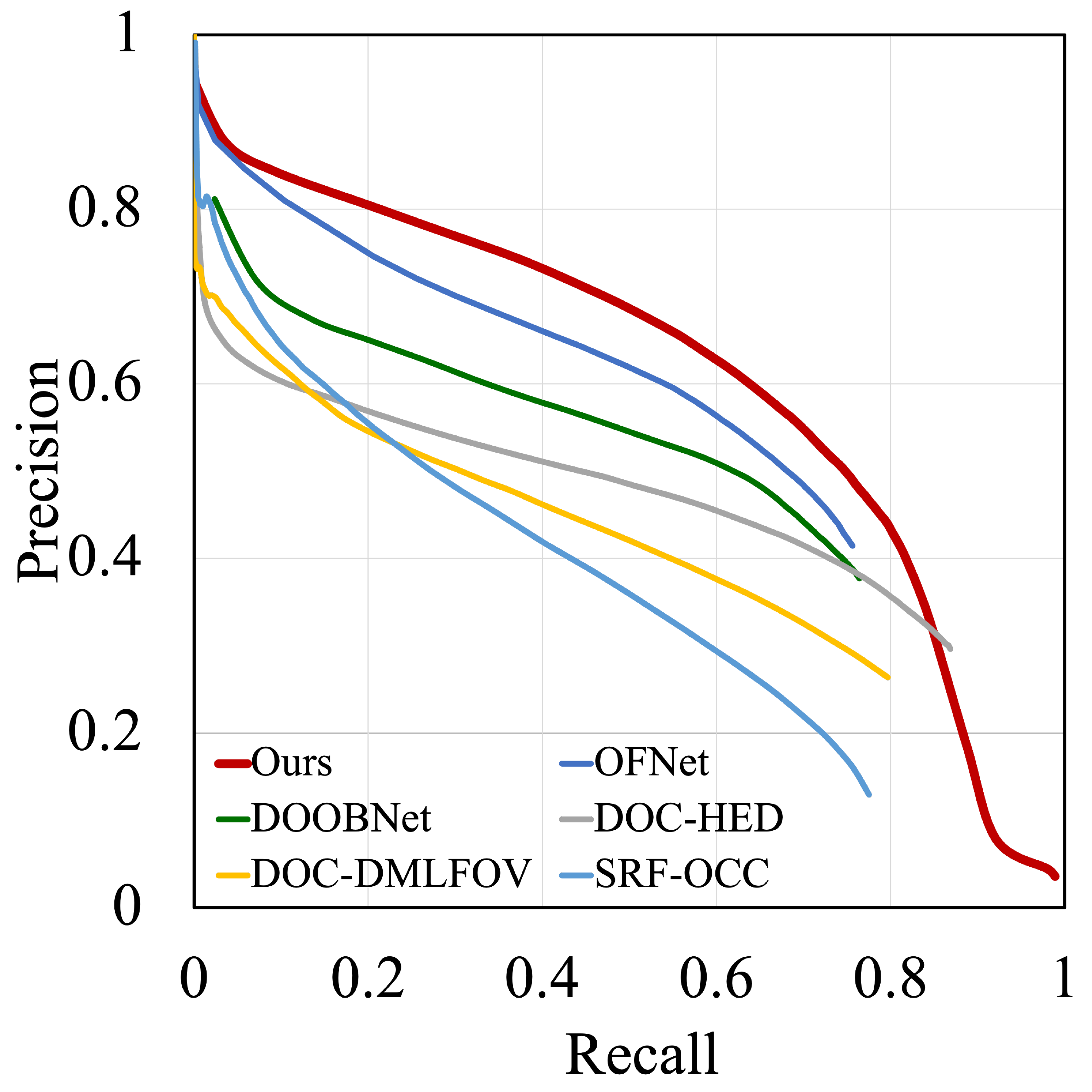} }
		\vspace*{-4mm}
		\caption{ The precision-recall curves(PRC)  of our method and others on PIOD  and BSDS ownership dataset. 
		}
		\label{piod_pr}
	\end{center}
	\vspace*{-6mm}
\end{figure*}

\subsection{Datasets and Implementation Details}

We evaluate the proposed approach on two public datasets: PIOD \cite{wang2016doc} and BSDS ownership dataset \cite{Figure_Ground_Assignment_in_Natural_Images}.

\noindent 
\textbf{PIOD dataset} contains 9,175 images for training and 925 images for testing. Each image is annotated with groundtruth object instance boundary map and corresponding orientation map.

\noindent
\textbf{BSDS ownership dataset} includes 100 training images and 100 testing images of natural scenes. Each image is annotated with groundtruth object instance boundary map and corresponding orientation map.

\noindent 
\textbf{Implementation details:} Our network is implemented in PyTorch. We use a ResNet50~\cite{resnet_2016,he2019bag} architecture pretrained on ImageNet as the backbone encoder.
AdamW optimizer is adopted to train our network. 
The learning rate is 3e-6.
Training images are randomly cropped into size $320 \times 320$ and form mini-batches of size 8.
We train our network on the PIOD dataset for 40k iterations and set hyperparameter $w_{\text{s-[2-5]}}$, $w_{\text{s-1}}$, $w_{\text{O}}$, $w_{\text{f}}$ and $\lambda$ to 0.5, 1.1, 1.1, 2.1, 1.7, respectively.
we train our network on the BSDS ownership dataset for 20k iterations and set $\lambda$ as 1.1. Other hyper-parameters are kept the same as in PIOD.

\noindent 
\textbf{Evaluation metrics:} We compute the precision and recall (\textbf{PR}) for boundary extraction  (\textbf{BPR})  and occlusion orientation (\textbf{OPR}). Three standard evaluation metrics are then calculated from PR: fixed contour threshold (\textbf{ODS}), best threshold of image (\textbf{OIS}) and average precision (\textbf{AP}). In the following sections, we use \textbf{B-Metric} to represent metrics calculated from BPR and  \textbf{O-Metric} to represent metrics calculated from OPR.
Note that, OPR is only calculated at the correctly detected boundary pixels.
																																																																										
\subsection{Comparison with the state-of-the-art methods}\label{ComparisonOtherWorks}
																																																																										
We compare our approach with recent methods including  SRF-OCC \cite{teo2015fast}, DOC-HED \cite{wang2016doc}, DOC-DMLFOV \cite{wang2016doc}, DOOBNet \cite{wang2018doobnet}, OFNet \cite{lu2019occlusion}.

\noindent \textbf{Performance on PIOD dataset:} The Precision-Recall curves of both boundary extraction and occlusion orientation tasks on PIOD dataset are shown in Figure \ref{piod_pr} (a) and (b). Clear improvements upon OFNet can be observed on both tasks. Table \ref{result_table} shows that even though trained alone on each subtask, our method still outperforms others by a large margin($3.5\%$ B-ODS and $4.3\%$ O-ODS).

\noindent \textbf{Performance on BSDS ownership dataset:}
The BSDS ownership dataset is difficult to train due to the small number of training samples. Nonetheless, our method can still achieve superior performance to others in both BPR and OPR as shown in Figure \ref{piod_pr}(c)(d). 
As shown in Table~\ref{res_table}, benefited from our proposed OOR, our method achieves a huge 10\% O-AP gain and an 8.3\% B-AP gain compared to state-of-the-art, showing strong potential in both orientation prediction and boundary extraction.

By substituting OOR with dor/dbr (OPNet+dor/dbr in Table~\ref{result_table}), OPNet with joint training strategy outperforms previous works but underperforms even the single boundary branch training of OPNet (Ours $\dag$), which verifies the superiority of our network structure and OOR, respectively.

\begin{figure*}[htbp]
	\begin{center}
		\setlength{\fboxsep}{0pt}
		\fcolorbox{white}{white}{%
			\begin{tabular}{p{0.001\linewidth}}\centering{}\end{tabular}%
		}
		\fcolorbox{white}{white}{%
			\begin{tabular}{p{0.085\linewidth}}\centering{GT}\end{tabular}%
		}
		\fcolorbox{white}{white}{%
			\begin{tabular}{p{0.085\linewidth}}\centering{\bf{Ours}}\end{tabular}%
		}
		\fcolorbox{white}{white}{%
			\begin{tabular}{p{0.085\linewidth}}\centering{DOOBNet}\end{tabular}%
		}
		\fcolorbox{white}{white}{%
			\begin{tabular}{p{0.085\linewidth}}\centering{OFNet}\end{tabular}%
		}
		\fcolorbox{white}{white}{%
			\begin{tabular}{p{0.085\linewidth}}\centering{GT}\end{tabular}%
		}
		\fcolorbox{white}{white}{%
			\begin{tabular}{p{0.085\linewidth}}\centering{\bf{Ours}}\end{tabular}%
		}
		\fcolorbox{white}{white}{%
			\begin{tabular}{p{0.085\linewidth}}\centering{DOOBNet}\end{tabular}%
		}
		\fcolorbox{white}{white}{%
			\begin{tabular}{p{0.085\linewidth}}\centering{OFNet}\end{tabular}%
		}
		\vfill
		\rotatebox{90}{\fcolorbox{white}{white}{%
			\textcolor{darkred}{\bf{OR}}
		}}
		\framebox{\includegraphics[width=0.109\linewidth]{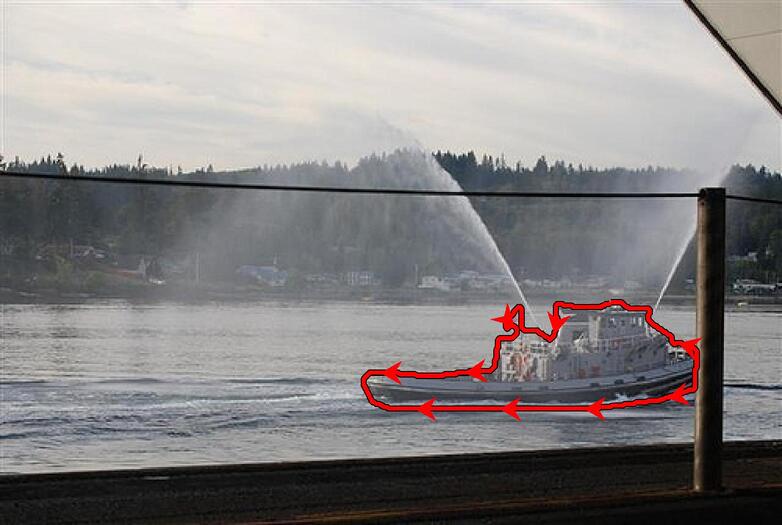}}
		\framebox{\includegraphics[width=0.109\linewidth]{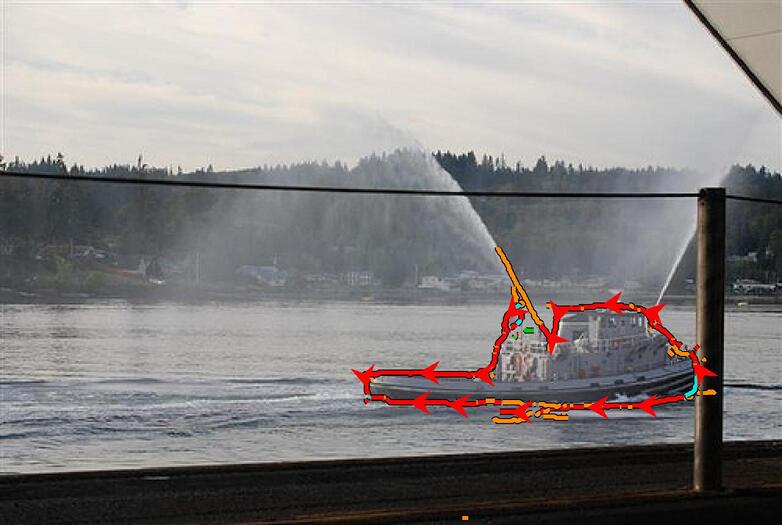}}
		\framebox{\includegraphics[width=0.109\linewidth]{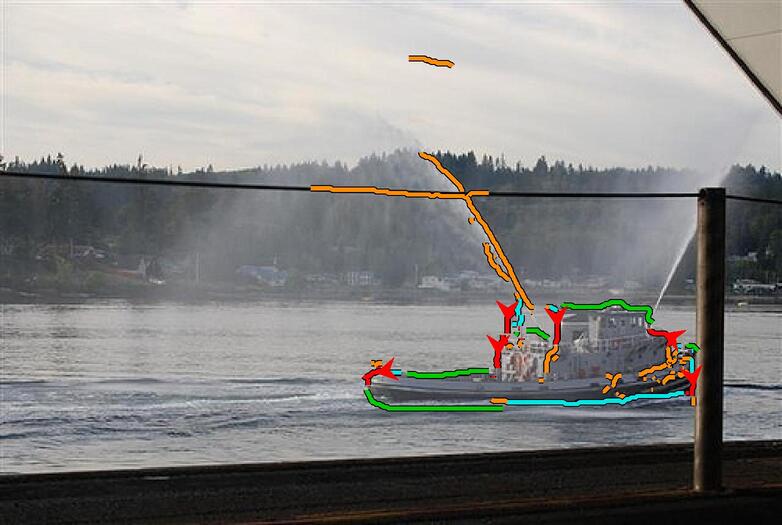}}
		\framebox{\includegraphics[width=0.109\linewidth]{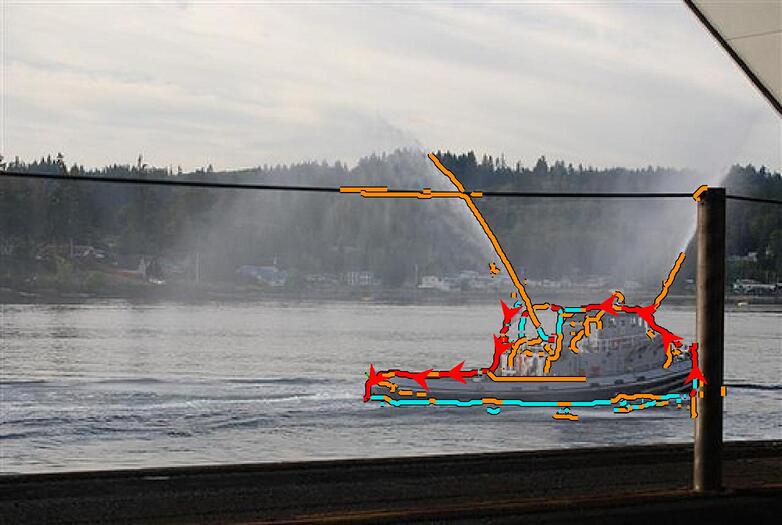}}
		\framebox{\includegraphics[width=0.109\linewidth]{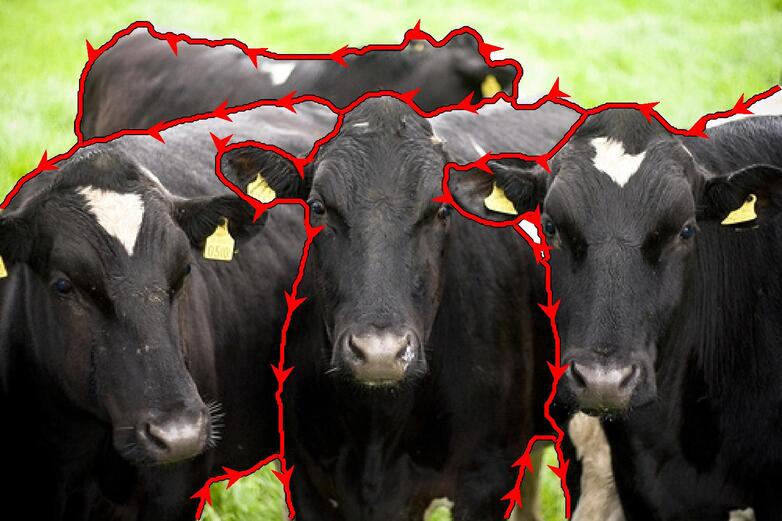}}
		\framebox{\includegraphics[width=0.109\linewidth]{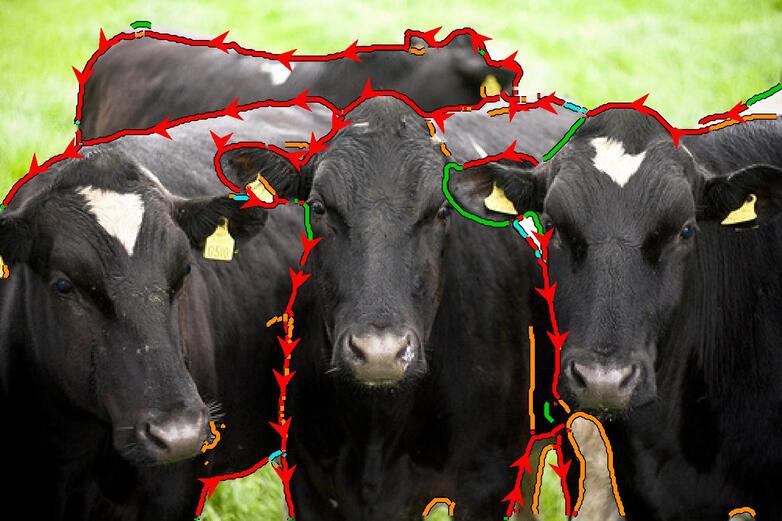}}
		\framebox{\includegraphics[width=0.109\linewidth]{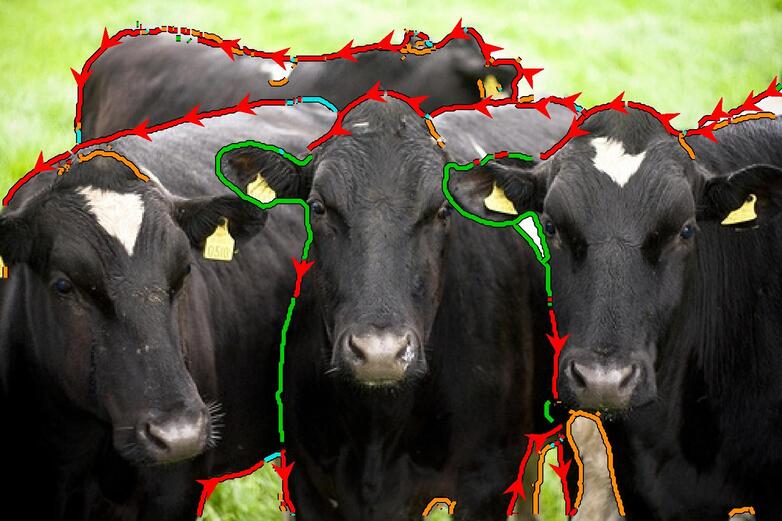}}
		\framebox{\includegraphics[width=0.109\linewidth]{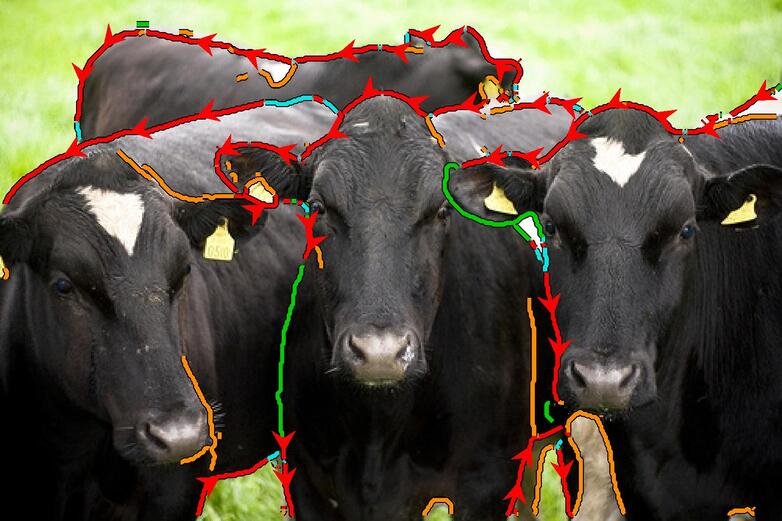}}
		\vfill
		\rotatebox{90}{\fcolorbox{white}{white}{%
			\textcolor{darkgreen}{\bf{BD}}
		}}
		\framebox{\includegraphics[width=0.109\linewidth]{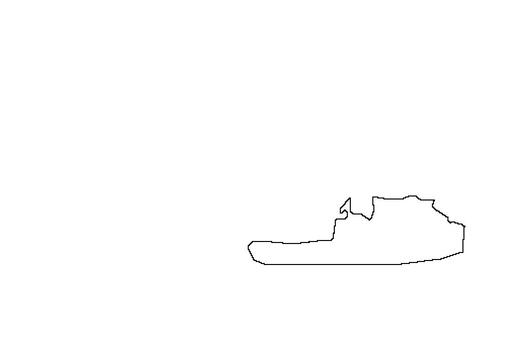}}
		\framebox{\includegraphics[width=0.109\linewidth]{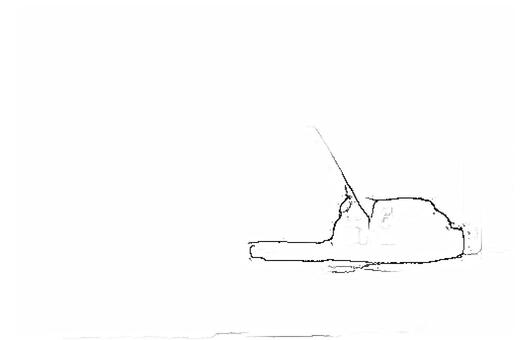}}
		\framebox{\includegraphics[width=0.109\linewidth]{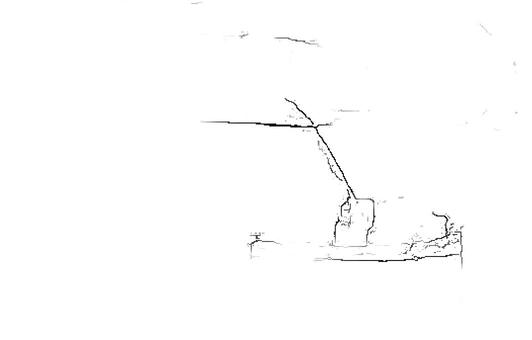}}
		\framebox{\includegraphics[width=0.109\linewidth]{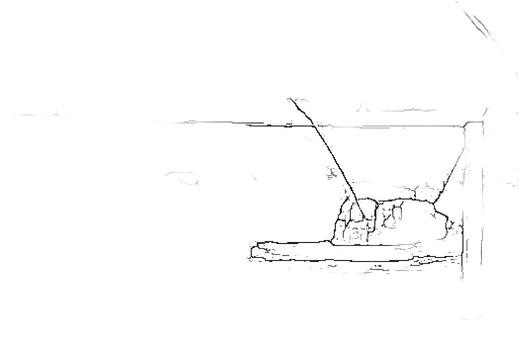}}
		\framebox{\includegraphics[width=0.109\linewidth]{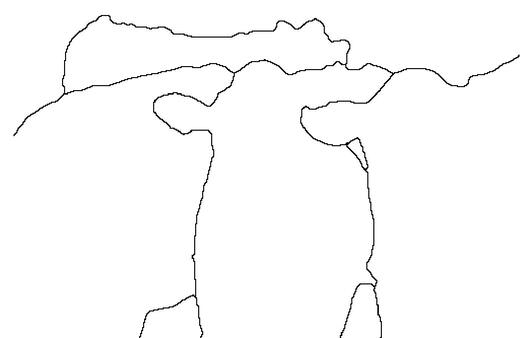}}
		\framebox{\includegraphics[width=0.109\linewidth]{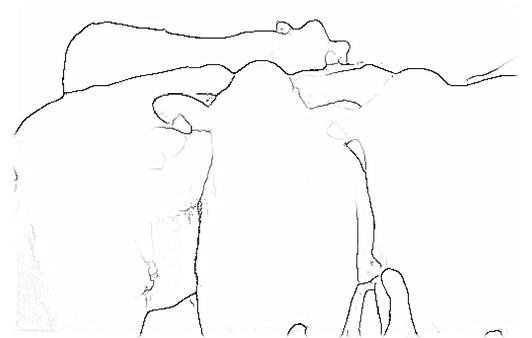}}
		\framebox{\includegraphics[width=0.109\linewidth]{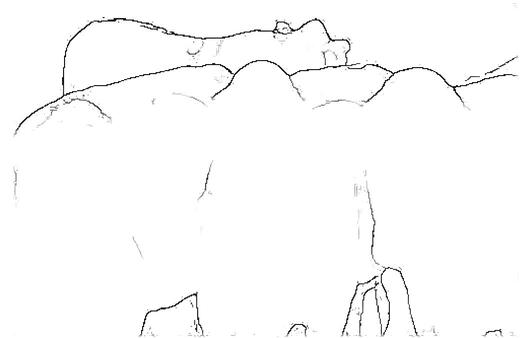}}
		\framebox{\includegraphics[width=0.109\linewidth]{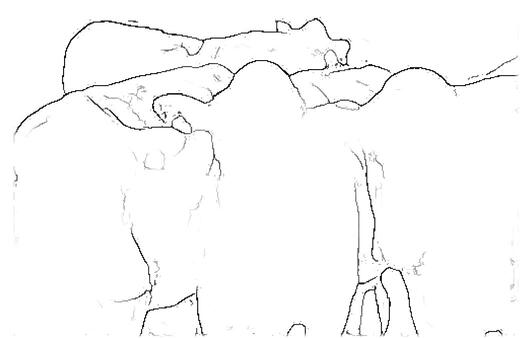}}
		\vfill
		\rotatebox{90}{\fcolorbox{white}{white}{%
			\textcolor{darkred}{\bf{OR}}
		}}
		\framebox{\includegraphics[width=0.109\linewidth]{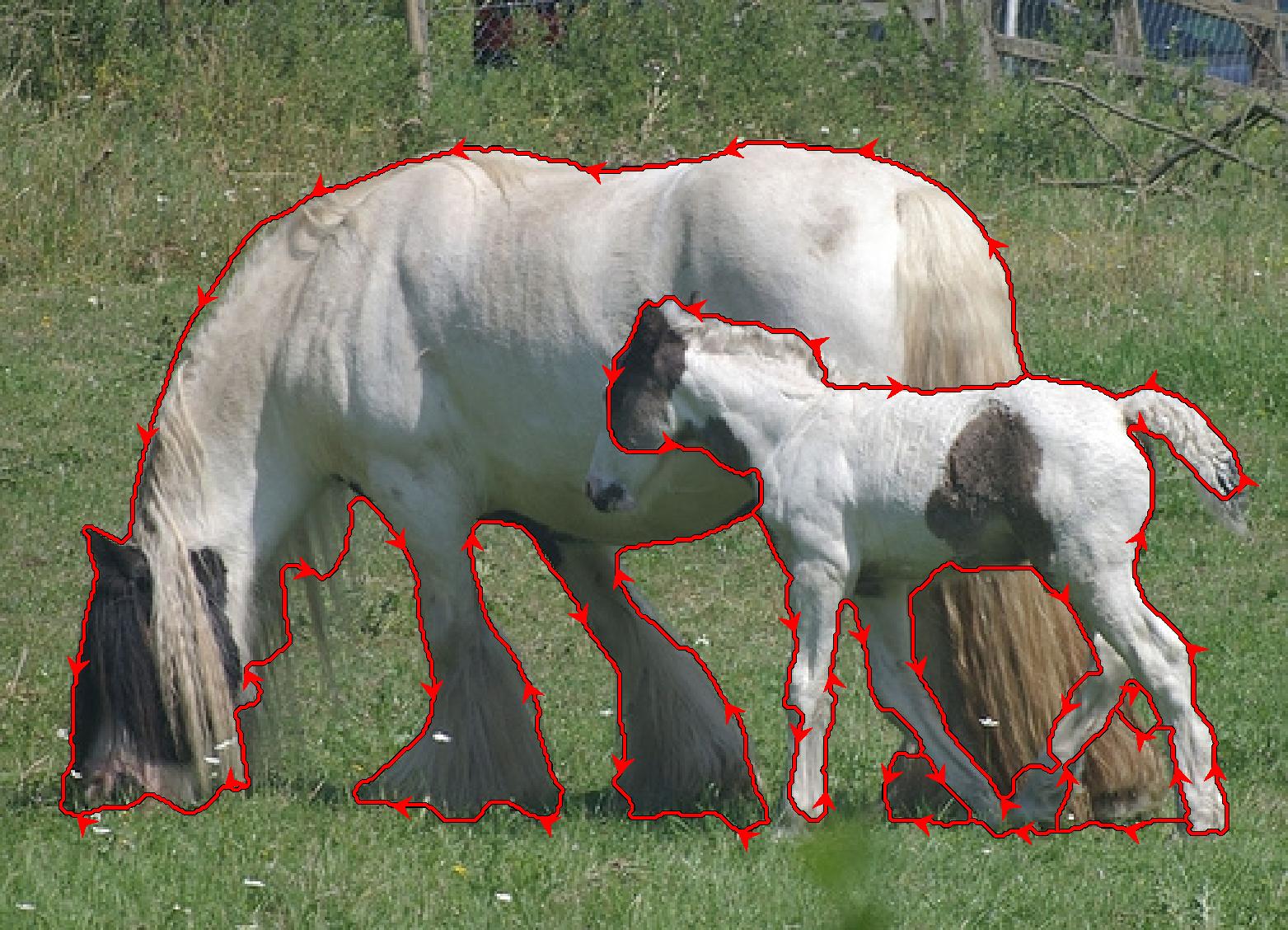}}
		\framebox{\includegraphics[width=0.109\linewidth]{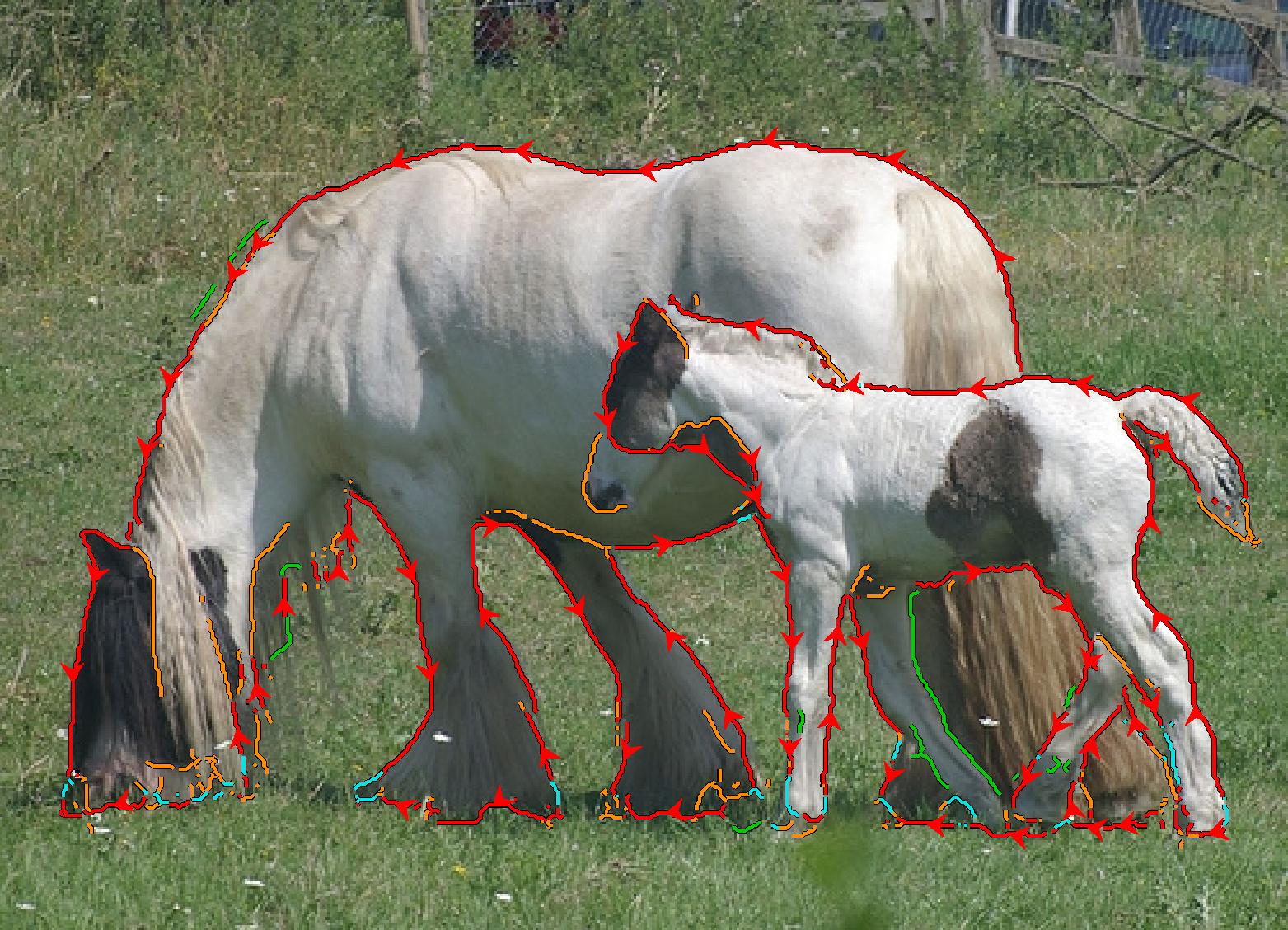}}
		\framebox{\includegraphics[width=0.109\linewidth]{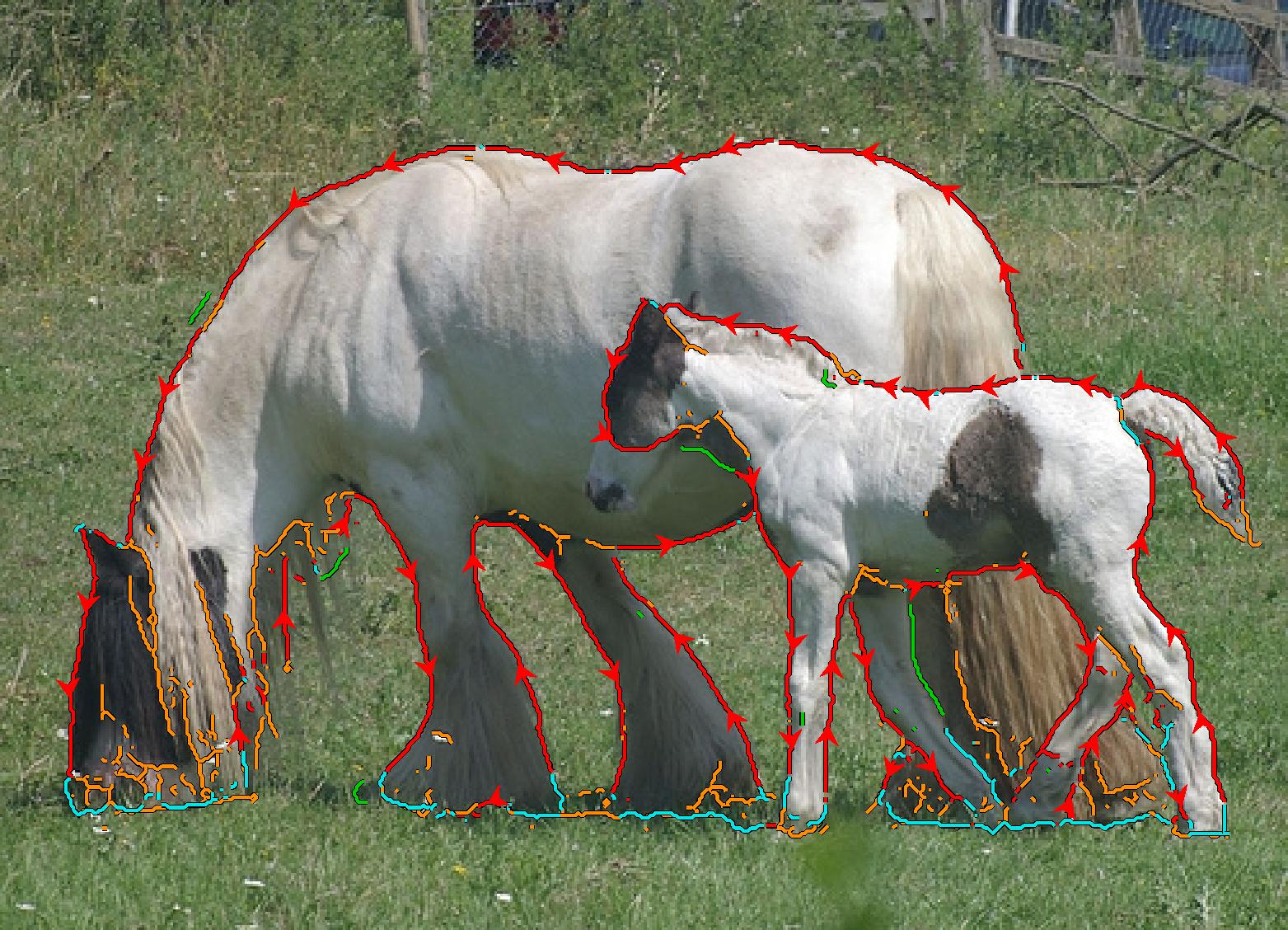}}
		\framebox{\includegraphics[width=0.109\linewidth]{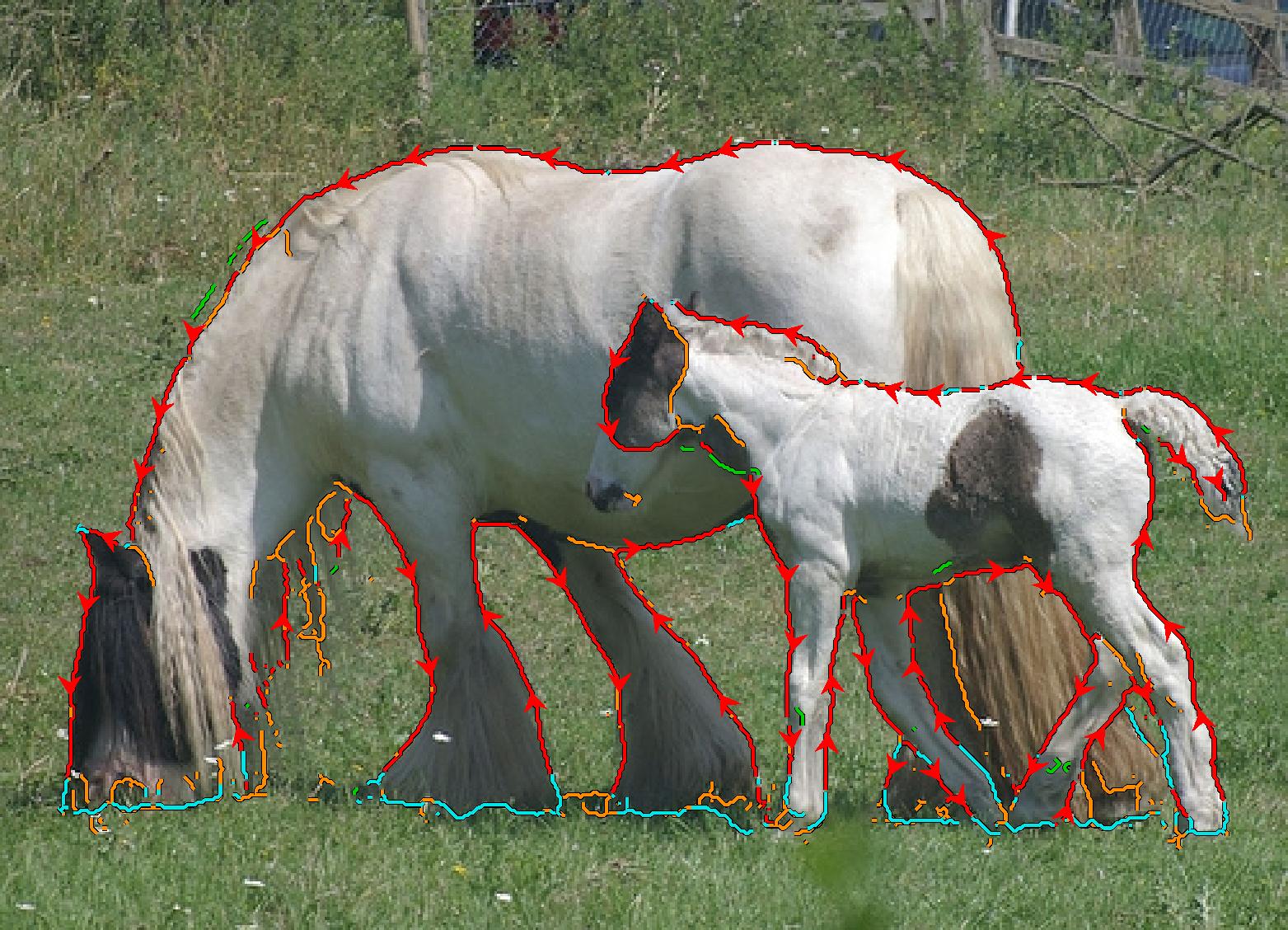}}
		\framebox{\includegraphics[width=0.109\linewidth]{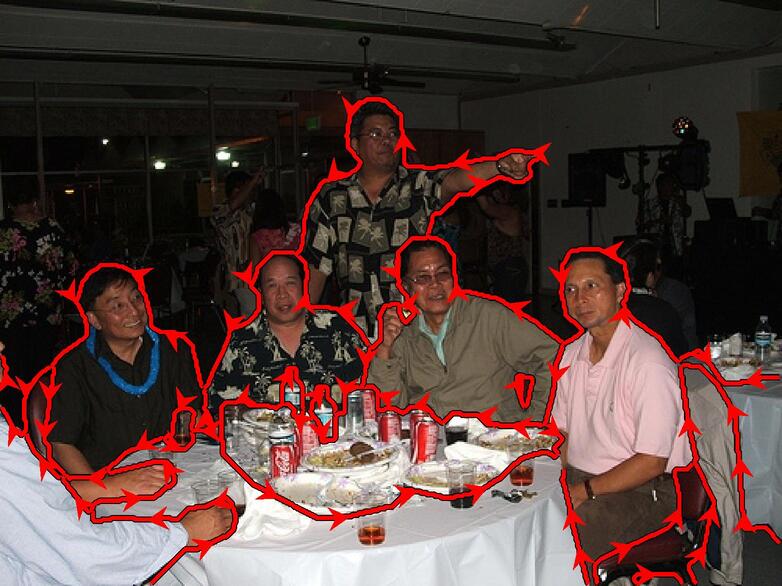}}
		\framebox{\includegraphics[width=0.109\linewidth]{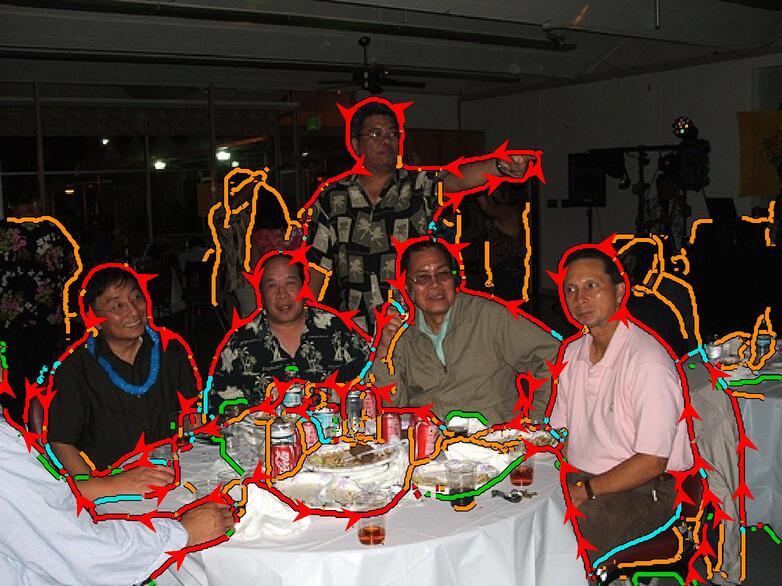}}
		\framebox{\includegraphics[width=0.109\linewidth]{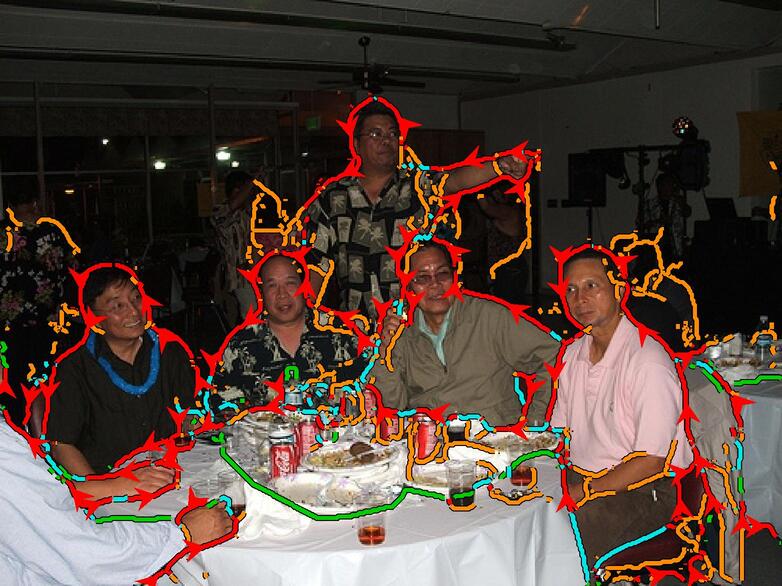}}
		\framebox{\includegraphics[width=0.109\linewidth]{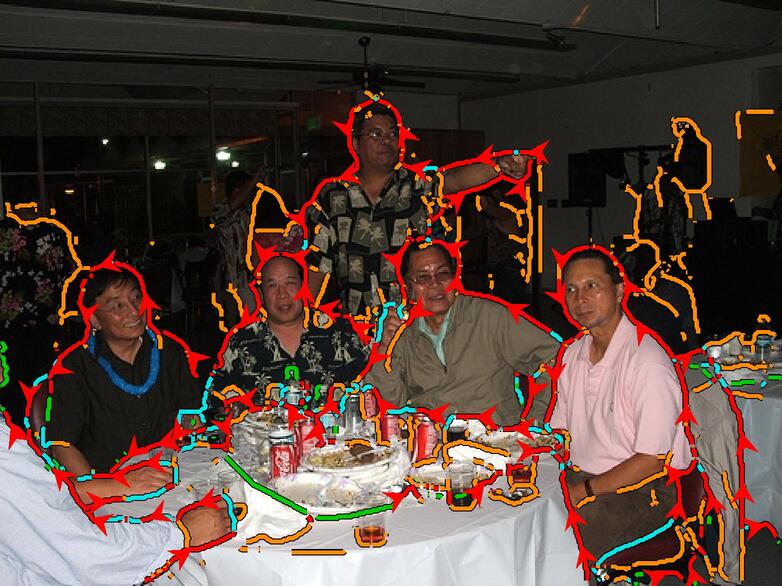}}
		\vfill
		\rotatebox{90}{\fcolorbox{white}{white}{%
			\textcolor{darkgreen}{\bf{BD}}
		}}
		\framebox{\includegraphics[width=0.109\linewidth]{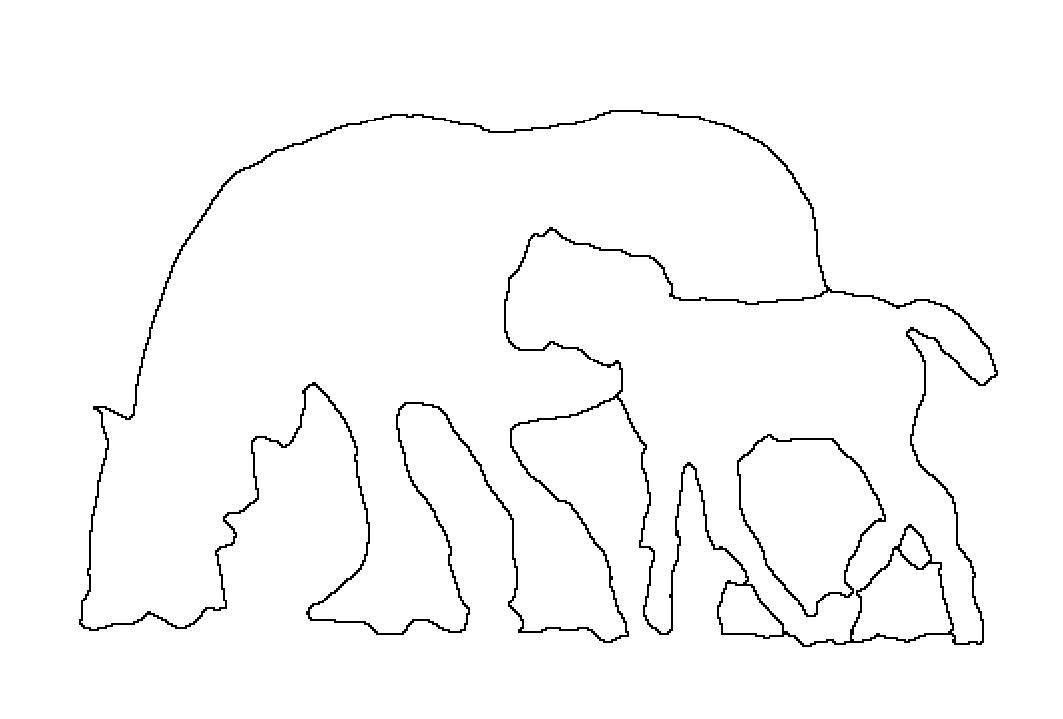}}
		\framebox{\includegraphics[width=0.109\linewidth]{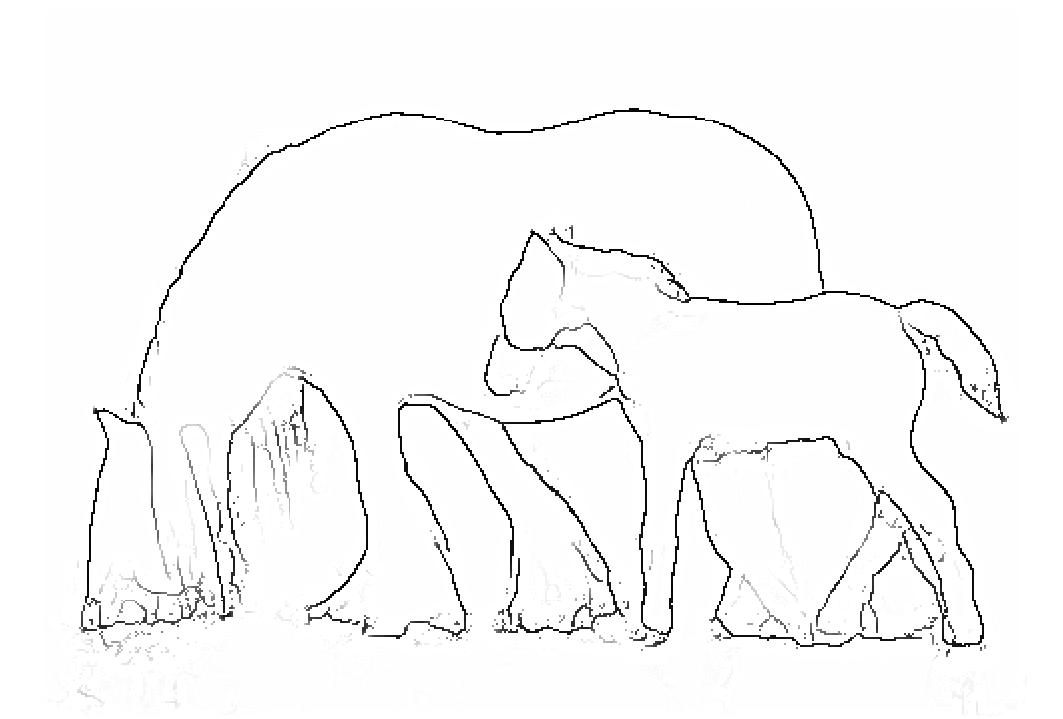}}
		\framebox{\includegraphics[width=0.109\linewidth]{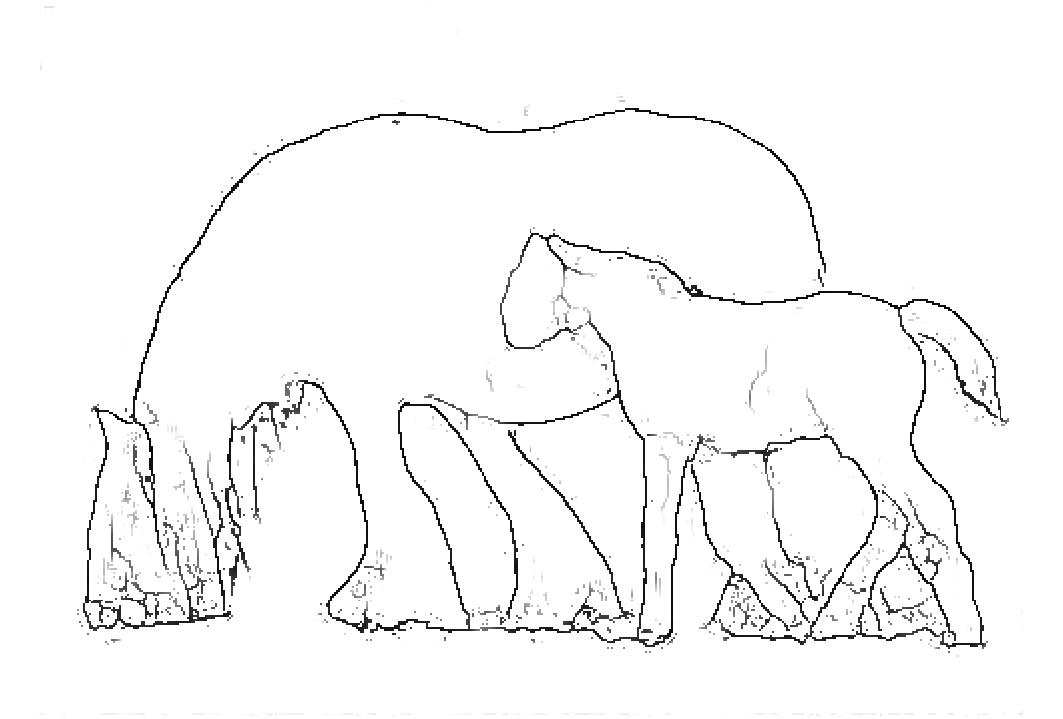}}
		\framebox{\includegraphics[width=0.109\linewidth]{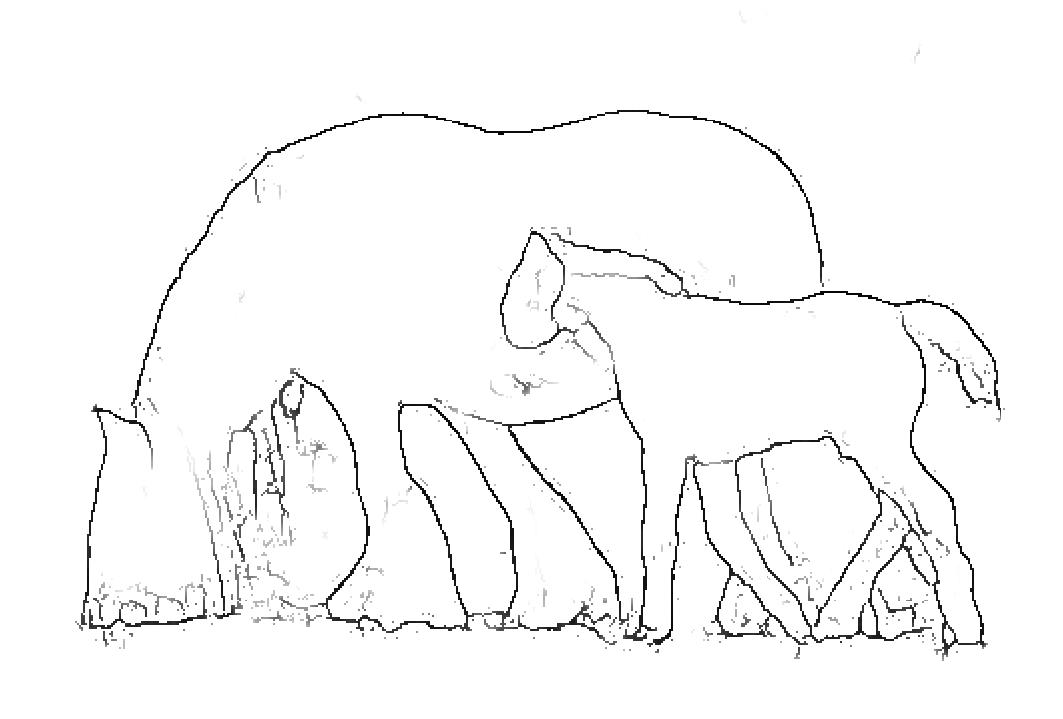}}
		\framebox{\includegraphics[width=0.109\linewidth]{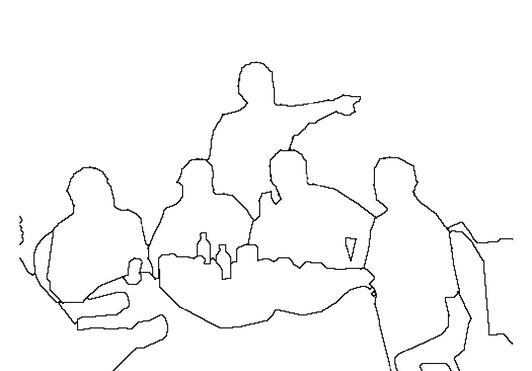}}
		\framebox{\includegraphics[width=0.109\linewidth]{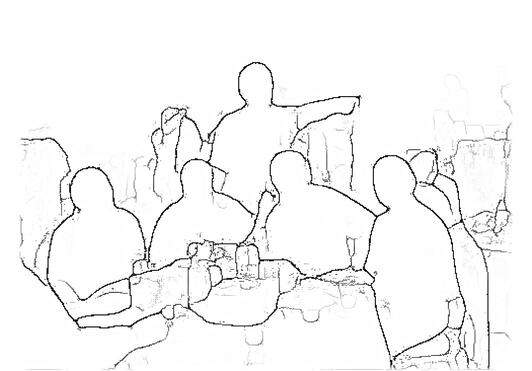}}
		\framebox{\includegraphics[width=0.109\linewidth]{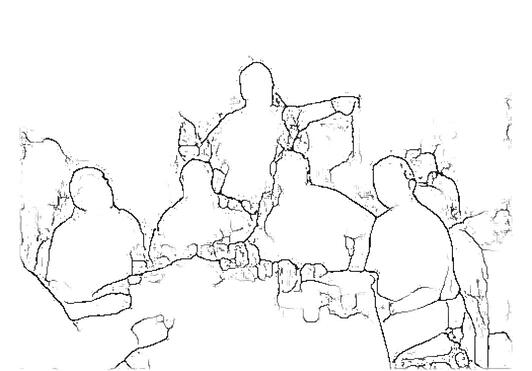}}
		\framebox{\includegraphics[width=0.109\linewidth]{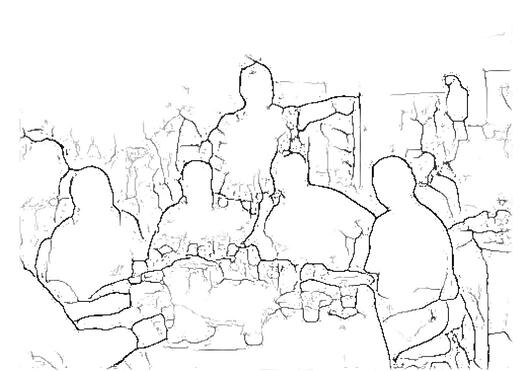}}
																																		
		\vfill
		\rotatebox{90}{\fcolorbox{white}{white}{%
			\textcolor{darkred}{\bf{OR}}
		}}
		\framebox{\includegraphics[width=0.109\linewidth]{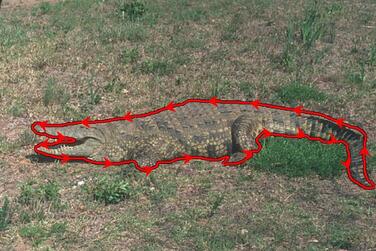}}
		\framebox{\includegraphics[width=0.109\linewidth]{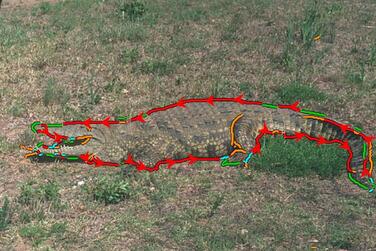}}
		\framebox{\includegraphics[width=0.109\linewidth]{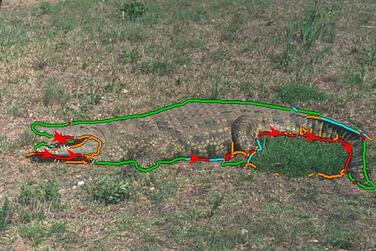}}
		\framebox{\includegraphics[width=0.109\linewidth]{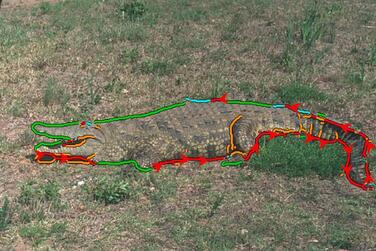}}
		\framebox{\includegraphics[width=0.109\linewidth]{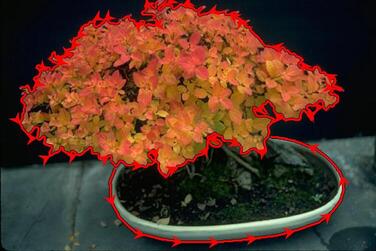}}
		\framebox{\includegraphics[width=0.109\linewidth]{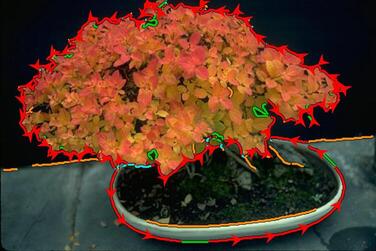}}
		\framebox{\includegraphics[width=0.109\linewidth]{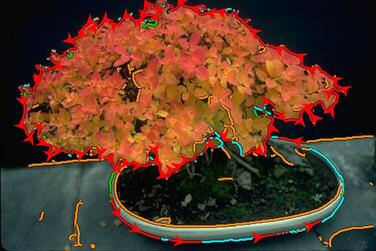}}
		\framebox{\includegraphics[width=0.109\linewidth]{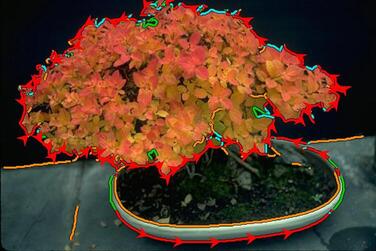}}
		\vfill
		\rotatebox{90}{\fcolorbox{white}{white}{%
			\textcolor{darkgreen}{\bf{BD}}
		}}
		\framebox{\includegraphics[width=0.109\linewidth]{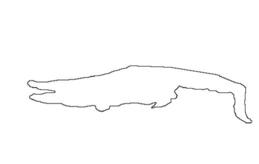}}
		\framebox{\includegraphics[width=0.109\linewidth]{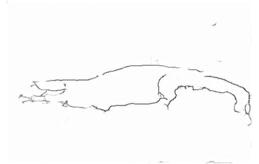}}
		\framebox{\includegraphics[width=0.109\linewidth]{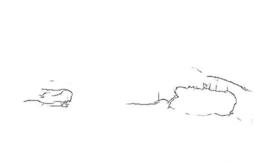}}
		\framebox{\includegraphics[width=0.109\linewidth]{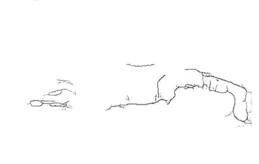}}
		\framebox{\includegraphics[width=0.109\linewidth]{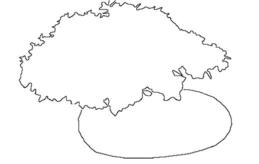}}
		\framebox{\includegraphics[width=0.109\linewidth]{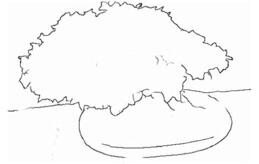}}
		\framebox{\includegraphics[width=0.109\linewidth]{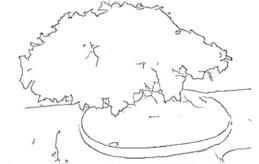}}
		\framebox{\includegraphics[width=0.109\linewidth]{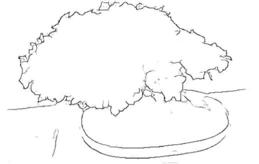}}
	\end{center}
	\vspace*{-4mm}
	\caption{Occlusion reasoning(\textcolor{darkred}{\textbf{OR}}) map and boundary(\textcolor{darkgreen}{\bf{BD}}) map predicted on PIOD (first four rows)  and BSDS ownership datasets (last two rows). We compare our method with DOOBNet and OFNet. GT refers to groundtruth map.
		{\color{red}red} pixels with arrows: correctly labeled occlusion boundaries; {\color{cyan}cyan}: correctly labeled boundaries but mislabeled occlusion; {\color{green}green}:
		false negative boundaries; {\color{orange}orange}: false positive boundaries. Our prediction results are visually much more accurate compared to DOOBNet and OFNet (Best viewed in color).}
	\label{res_img}
	\vspace*{-4mm}
\end{figure*}

Finally, we visualize the boundary maps and occlusion relationship maps, as shown in Figure \ref{res_img}. 
It can be observed that our method extracts more complete and clearer boundaries on multiple images than DOOBNet\cite{wang2018doobnet} and OFNet\cite{lu2019occlusion}.
In complex and indistinguishable scenes, such as images on the right-hand side, we are still ahead of others, showing stronger generalization potential.
As shown in the last image sets on the left-hand side, our model can even discern the leg which is not labeled on the groundtruth map.
In summary, our method can outperform others by a large margin both in boundary extraction and occlusion orientation prediction, which validates its effectiveness.

\begin{table}[ht]
	\renewcommand\arraystretch{1.1}
	\begin{center}
		\caption{Comparison on number of shared layers and multi-scale supervision on PIOD.}
		\label{shard_layer_table}
		\scalebox{0.93}{
			\setlength{\tabcolsep}{1.4mm}{
				\begin{tabular}
					{c | c | c c c c c c}
					\toprule[1.2pt]
					\multicolumn{2}{c}{\multirow{2}{*}{~}}
					&\multicolumn{3}{|c}{\bf{BPR}}&\multicolumn{3}{c}{\bf{OPR}}\cr
					\cmidrule(r){3-5} \cmidrule(r){6-8}
					\multicolumn{2}{c|}{~}
					&\bf{ODS}&\bf{OIS}&\bf{AP}&\bf{ODS}&\bf{OIS}&\bf{AP}\cr
					\midrule[1pt]
					\multirow{4}{*}{
					\makecell[c]{\bf{Layer} \leavevmode \\ \bf{Shared}}
					}
					  & top-1 & $78.7$         & $79.8$        & $80.8$        & $76.2$        & $77.1$        & $77.1$\cr        
					  & top-2 & { $\bf{79.5}$} & {$\bf{80.4}$} & {$\bf{83.1}$} & {$\bf{77.1}$} & {$\bf{77.8}$} & {$\bf{79.4}$}\cr 
					  & top-3 & $77.6$         & $78.6$        & $80.6$        & $75.4$        & $76.3$        & $77.5$\cr        
					  & top-4 & $77.0$         & $78.4$        & $79.7$        & $74.8$        & $76.0$        & $76.4$\cr        
					\midrule[0.8pt]
					\multirow{5}{*}{
					\makecell[c]{\bf{Number} \leavevmode \\ \bf{of} \leavevmode \\ \bf{Scales}}
					}
					  & 1     & $75.4$         & $76.5$        & $77.4$        & $73.1$        & $74.0$        & $74.1$\cr        
					  & 2     & $78.6$         & $79.5$        & $82.6$        & $76.3$        & $76.4$        & $78.9$\cr        
					  & 3     & $78.1$         & $79.8$        & $82.0$        & $76.4$        & $77.3$        & $78.6$\cr        
					  & 4     & $78.6$         & $79.5$        & $81.1$        & $76.1$        & $76.9$        & $77.6$\cr        
					  & 5     & $\bf{79.5}$    & $\bf{80.4}$   & $\bf{83.1}$   & $\bf{77.1}$   & $\bf{77.8}$   & $\bf{79.4}$\cr   
					\bottomrule[0.8pt]
				\end{tabular}}}
	\end{center}
	\vspace*{-4mm}
\end{table}
			
\subsection{Ablation Study}\label{AblationStudy}

In this section, we conduct experiments on the PIOD dataset to study the impact of different architectural designs and verify each component in our network.

\textbf{Comparison on number of shared layers: } 
We compared sharing top-1, top-2, top-3, and top-4 stages at high stages. As shown in Table~\ref{shard_layer_table}, the best result appears when sharing the top-2 stages, surpassing the second best(top-1) by 2.3\% in B-AP. When the sharing extends to the lower stages, the result begins to deteriorate, which empirically proves our argument that the two branches should share occlusion information existing in higher stages and remain separated in lower stages to recover task-specific spatial information.

\textbf{Comparison of orientation losses: } 
We change the output of our occlusion orientation path so that it can regress the continuous orientation value, as in previous work.
We then compare the performance of predicting continuous orientation value(DOOBNet) and predicting orthogonal vectors(OOR) with our network structure.
As shown in Table \ref{res_table}, the proposed OOR and OOR loss surpasses DOOBNet's orientation representation by 6.5\% and 10\% O-AP on PIOD and BSDS ownership dataset, respectively, which validates the effectiveness of our orthogonal orientation representation and loss.

\textbf{Multi-scale supervision effect:}
Our boundary path outputs five side boundary maps with multi-scale supervision signals. To validate the effectiveness of such multi-scale supervision, we compared results of supervising the lowest one to the lowest five side boundary maps, where a lower boundary map has a larger spatial size.
As shown in Table \ref{shard_layer_table}, 
the performance of only supervising the lowest side is significantly worse than other settings.
Performance is also greatly improved by using five-scale supervision, surpassing the second best(using 4 side maps) by 2.0\% B-AP, which demonstrates the importance of multi-scale supervision and multi-result fusion.

\section{Conclusion}

In this paper, we present the novel OPNet, which shares occlusion features in the deep stage and splits into two separate decoder paths with larger spatial sizes. Besides, we apply multi-scale supervision to the boundary extraction. We also present a robust orthogonal occlusion representation, which avoids the endpoint prediction errors and negative impact of angle periodicity. Our network achieves new state-of-the-art performances on both PIOD and BSDS ownership datasets.

\section*{Acknowledgement}

This work was supported by the national key R \& D program intergovernmental international science and technology innovation cooperation project (2021YFE0101600).

{\small
	\bibliographystyle{ieee_fullname}
	\bibliography{egbib}
}

\end{document}